%% file: paper.tex
\documentclass[10pt,journal,compsoc]{IEEEtran}

\ifCLASSOPTIONcompsoc
  \usepackage[nocompress]{cite}
\else
  \usepackage{cite}
\fi

\input{math_commands.tex}

\usepackage{bbding}
\usepackage{pifont}
\usepackage{wasysym}
\usepackage{amssymb}
\usepackage{float}
\usepackage{dblfloatfix}
\usepackage{footnote}
\makesavenoteenv{tabular}
\makesavenoteenv{table}
\makesavenoteenv{figure}
\usepackage{authblk}

\usepackage{hyperref}
\usepackage{gensymb}
\usepackage{url}
\usepackage{xspace}
\usepackage{caption}
\usepackage{mathtools}
\usepackage{amsfonts}
\usepackage{booktabs}
\usepackage{graphicx}
\usepackage{soul}
\usepackage{color}
\usepackage{xspace}
\usepackage[subrefformat=parens]{subfig}
\usepackage{multicol}

\def\etal{\emph{et al}.}

\begin{document}
\title{Pruning Convolutional Neural Networks\\
with Self-Supervision}

\author[1,2]{Mathilde Caron}
\author[1]{Ari Morcos}
\author[1]{Piotr Bojanowski}
\author[2]{Julien Mairal}
\author[1]{Armand Joulin}
\affil[1]{Facebook AI Research}
\affil[2]{Univ. Grenoble Alpes, Inria, CNRS, Grenoble INP, LJK, 38000 Grenoble, France}


\IEEEtitleabstractindextext{%
\begin{abstract}
Convolutional neural networks trained without supervision come close to matching performance with supervised pre-training, but sometimes at the cost of an even higher number of parameters.
Extracting subnetworks from these large unsupervised convnets with preserved performance is of particular interest to make them less computationally intensive.
Typical pruning methods operate during training on a task while trying to maintain the performance of the pruned network on the same task.
However, in self-supervised feature learning, the training objective is agnostic on the representation transferability to downstream tasks.
Thus, preserving performance for this objective does not ensure that the pruned subnetwork remains effective for solving downstream tasks. 
In this work, we investigate the use of standard pruning methods, developed primarily for supervised learning, for networks trained without labels (i.e. on self-supervised tasks).	
We show that pruned masks obtained with or without labels reach comparable performance when re-trained on labels, suggesting that pruning operates similarly for self-supervised and supervised learning.
Interestingly, we also find that pruning preserves the transfer performance of self-supervised subnetwork representations.
\end{abstract}

}

\maketitle
\IEEEdisplaynontitleabstractindextext
\IEEEpeerreviewmaketitle

\input{introduction.tex}
\input{related_work.tex}
\input{approach.tex}

\input{experiments.tex}

\input{discussion.tex}

\section*{Acknowledgment}
We thank the members of Thoth and FAIR teams for their help and fruitful discussions.
Julien Mairal was funded by the ERC grant number 714381 (SOLARIS project).


\bibliography{egbib}
\bibliographystyle{IEEEtran}


\begin{IEEEbiography}[{\includegraphics[width=1in,height=1.25in,clip,keepaspectratio]{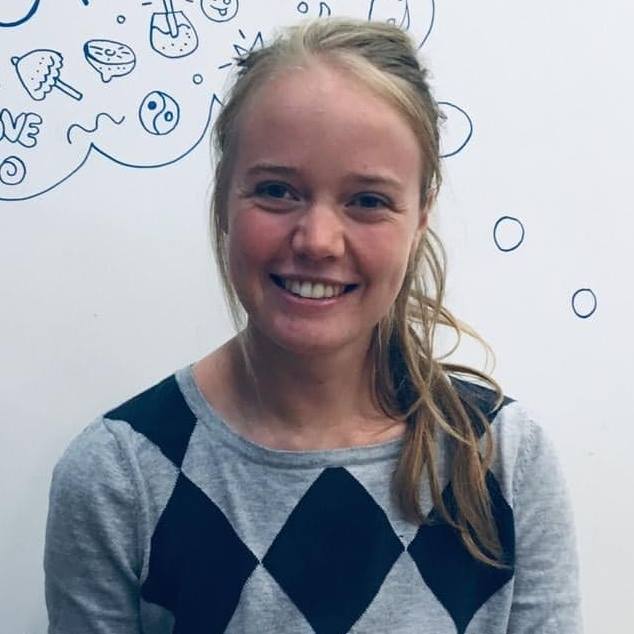}}]{Mathilde Caron}
is currently a second-year PhD student both at Inria in the Thoth team and at Facebook AI Research (FAIR) working on large-scale unsupervised representation learning for vision.
Her supervisors are Julien Mairal, Piotr Bojanowski and Armand Joulin.
Previously, she did her master thesis on clustering for unsupervised learning of visual representation at FAIR under the supervision of Piotr Bojanowski.
Before that, she graduated from both Ecole polytechnique and KTH Royal Institute of Technology where she was mostly interested in applied mathematics and statistical learning.
\end{IEEEbiography}

\vspace{-2.5em}

\begin{IEEEbiography}[{\includegraphics[width=1in,height=1.25in,clip,keepaspectratio]{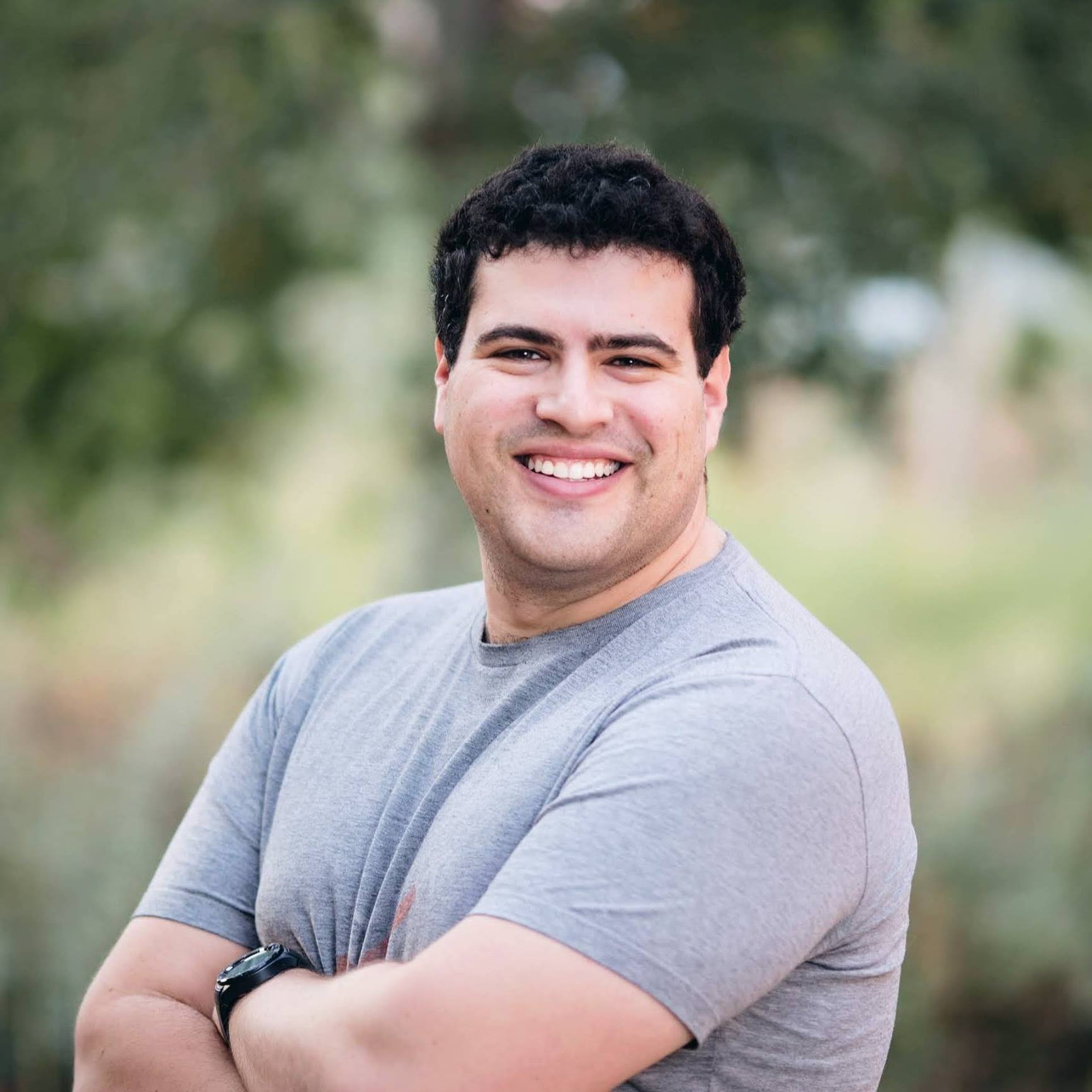}}]{Ari Morcos}
is a research scientist at Facebook AI Research (FAIR) in Menlo Park, working on using neuroscience-inspired approaches to understand and build better machine learning systems.
Previously, he worked at DeepMind in London.
He earned his PhD working with Chris Harvey at Harvard University.
For his thesis, he developed methods to understand how neuronal circuits perform the computations necessary for complex behavior.
In particular, his research focused on how parietal cortex contributes to evidence accumulation decision-making.
For his undergraduate work, he attended UCSD, where he worked with Fred Gage to investigate the role of REST/NRSF in adult neurogenesis.
\end{IEEEbiography}

\vspace{-2.5em}

\begin{IEEEbiography}[{\includegraphics[width=1in,height=1.25in,clip,keepaspectratio]{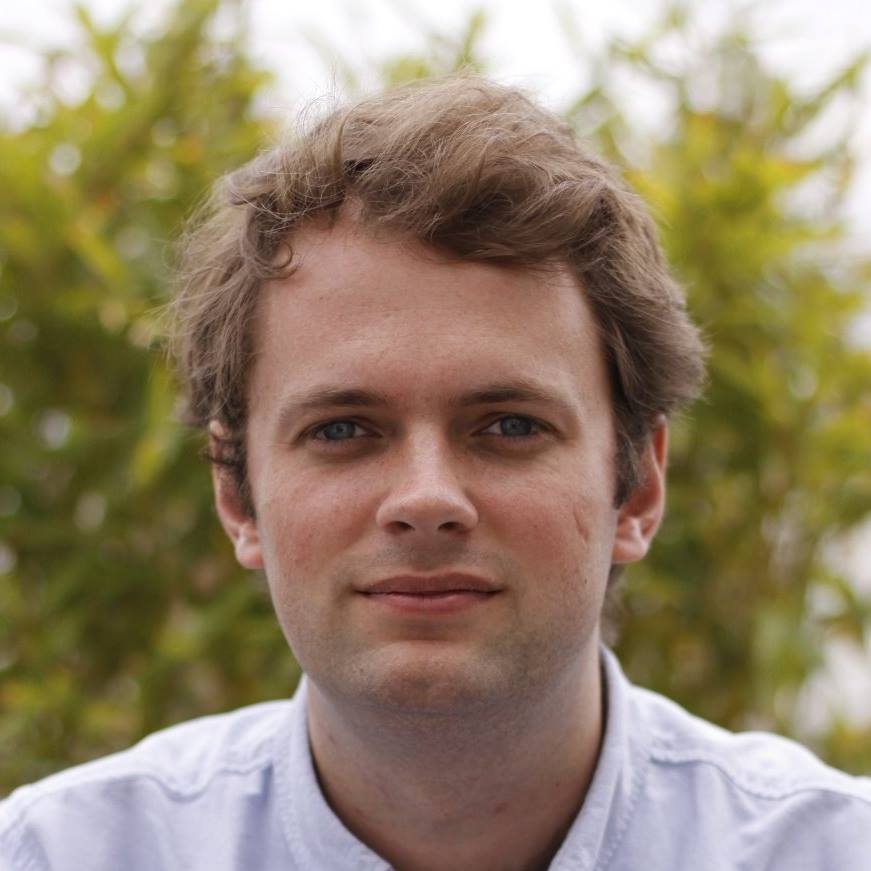}}]{Piotr Bojanowski}
is a research scientist at Facebook AI Research, working on machine learning applied to computer vision and natural language processing.
His main research interest revolve around large-scale unsupervised learning.
Before joining Facebook, in 2016, he got a PhD in Computer Science at the Willow team (INRIA Paris) under the supervision of Jean Ponce, Cordelia Schmid, Ivan Laptev and Josef Sivic.
He graduated from Ecole polytechnique in 2013 and received a Masters Degree in Mathematics, Machine Learning and Computer Vision (MVA).
\end{IEEEbiography}

\vspace{-2.5em}

\begin{IEEEbiography}[{\includegraphics[width=1in,height=1.25in,clip,keepaspectratio]{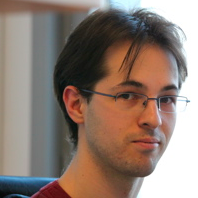}}]{Julien Mairal}
(SM16) received the Graduate degree from the Ecole Polytechnique, Palaiseau, France, in 2005, and the Ph.D. degree from
Ecole Normale Superieure, Cachan, France, in
2010. He was a Postdoctoral Researcher at the
Statistics Department, UC Berkeley. In 2012, he
joined Inria, Grenoble, France, where he is currently a Research Scientist. His research interests include machine learning, computer vision,
mathematical optimization, and statistical image
and signal processing. In 2016, he received a
Starting Grant from the European Research Council and in 2017, he
received the IEEE PAMI young research award. He was awarded the
Cor Baayen prize in 2013, the IEEE PAMI young research award in 2017
and the test-of-time award at ICML 2019.
\end{IEEEbiography}

\vspace{-2.5em}

\begin{IEEEbiography}[{\includegraphics[width=1in,height=1.25in,clip,keepaspectratio]{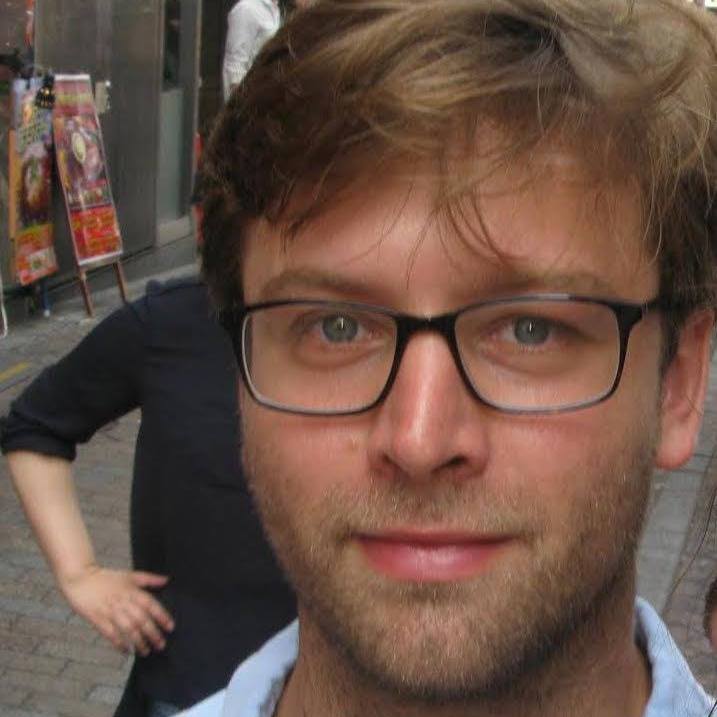}}]{Armand Joulin}
is a research manager at Facebook AI Research.
Prior to this position, he was a postdoctoral fellow at Stanford University working with Fei-Fei Li and Daphne Koller.
He did his PhD in Inria and Ecole Normale Superieure, under the supervision of Francis Bach and Jean Ponce.
He did his undergrad at Ecole Polytechnique.
His subjects of interest are machine learning, computer vision and natural language processing.
\end{IEEEbiography}

\newpage
\input{appendices.tex}

\end{document}

%% file: math_commands.tex
\usepackage{amsmath,amsfonts,bm}

\def\eqref#1{equation~\ref{#1}}

\def\1{\bm{1}}

\DeclareMathAlphabet{\mathsfit}{\encodingdefault}{\sfdefault}{m}{sl}
\SetMathAlphabet{\mathsfit}{bold}{\encodingdefault}{\sfdefault}{bx}{n}



%% file: introduction.tex
\IEEEraisesectionheading{\section{Introduction}\label{sec:introduction}}

\IEEEPARstart{C}{onvolutional} neural networks (convnets) pre-trained without supervision are emerging from the image recognition community with performance approaching that of supervised ImageNet pre-training~\cite{caron2019unsupervised,goyal2019scaling}.
They usually contain a huge number of parameters: some even count hundreds of millions of weights~\cite{bachman2019learning,henaff2019data} which is an order of magnitude bigger than standard networks used for supervised pre-training~\cite{he2016deep}.
Extracting subnetworks from these large unsupervised convnets with preserved representation power would make both training and use of their visual features less computationally intensive.
A well-established solution to reduce parameter-counts of large neural networks is to prune part of their weights~\cite{lecun1990optimal}.
However, pruning has been originally developed in a fully-supervised context: it is usually operated while or after training on a supervised task with the goal of preserving the validation accuracy of a subnetwork on the same task~\cite{louizos2017learning,han2015learning}.
Yet, in self-supervised learning, the task used to train a network is often merely a proxy~\cite{noroozi2016unsupervised,gidaris2018unsupervised}, thus preserving performance on this surrogate does not guarantee preserved transferability of the resulting subnetwork.
To the best of our knowledge, there is no study in the current literature on the impact of pruning on networks pre-trained with self-supervised tasks.
\\

For these reasons, we explore in this work if pruning methods that were developed for supervised learning can be used for networks trained from unlabeled data only.
The main questions about pruning networks trained with self-supervision we are trying to answer are the following:
How subnetworks obtained from self-supervision compare to subnetworks pruned with labels?
Can subnetworks pruned from unlabeled data be used for supervised tasks?
Does pruning networks pre-trained with self-supervision deteriorate the quality of their subsequent features when transferred to different downstream tasks?
\\

To investigate these questions, we propose a simple pipeline based on well-established methods from the unstructured pruning and self-supervised learning literatures.
In particular, we use the magnitude-based iterative pruning technique of Han~\etal~\cite{han2015learning}, which compresses networks by alternatively training with labels and pruning the network parameters with the smallest magnitude.
Recent works have shown that subnetworks uncovered by Han~\etal~\cite{han2015learning} lead to accuracy comparable with that of an unpruned network when re-trained from a selected inherited set of weights termed the ``winning tickets''~\cite{frankle2018lottery,frankle2019lottery} or even from random initialization though for moderate pruning rates~\cite{liu2018rethinking}.
In our work, we build upon these works and simply replace semantic labels by pseudo-labels given by self-supervision pretext tasks.
Interestingly, Morcos~\etal~\cite{morcos2019one} have shown that winning tickets initializations can be re-used across different datasets with a common domain (natural images) trained on the same task (labels classification).
In contrast, in this work, we explore among other things if winning tickets initializations from self-supervised tasks can be used as initialization for training label classification on a commun dataset.
\\

Our experiments show that pruning self-supervised networks with a standard pruning method preserves the resulting features quality when evaluated on different downstream tasks.
We also observe that transferring an already pruned pre-trained network gives better transfer performance than pruning a pre-trained network directly on the transfer objective.
This is convenient since the subsequent optimal scenario is to provide already pruned pre-trained networks, thus dispensing the need for users to operate any pruning on the target task.
Beside, we find that the pruned subnetworks obtained with self-supervision can be re-trained successfully on ImageNet labels classification, and are even on par with supervised pruning when randomly re-initialized.
More precisely, the quality of the pruned mask alone is similar between supervised and self-supervised pruning but the winning tickets initializations of self-supervised subnetworks are not as good starting points as the ones inherited from label classification task directly.
Overall, we find that pruning networks trained with self-supervision works well, in the sense that the transfer performance of the pruned subnetworks is preserved even for high pruning rates and they can be re-trained from scratch on labels.
As a matter of fact, we choose to conduct most of our experiments on ImageNet; we remark indeed that deep networks trained on smaller datasets such as CIFAR-10 are already sparse at convergence, making conclusions drawn about pruning potentially misleading if this effect is not accounted for.

%% file: related_work.tex
\section{Related work}
\label{sec:related_work}

\noindent\textbf{Pruning.} 
Pruning is an approach to model compression~\cite{han2015learning} and regularization~\cite{lecun1990optimal} in which weights or nodes/filters are removed, typically by clamping them to zero (see the work of Liu~\etal~\cite{liu2018rethinking} for a review of the different pruning methods).
It is an active research which has primarily focused on pruning an already trained network~\cite{han2015learning,li2016pruning} or pruning while training~\cite{prakash2019repr}.
In particular, the iterative pruning during training of Han~\etal~\cite{han2015learning} has been extended to continuous pruning~\cite{guo2016dynamic}, layer-wise pruning~\cite{dong2017learning} and with weight sharing~\cite{ullrich2017soft}.
Pruning during training has been considered with $\ell_0$ regularization~\cite{louizos2017learning}, binary convolution~\cite{rastegari2016xnor} or using the hashing trick for weight sharing~\cite{chen2015compressing}.
\\

\noindent\textbf{The lottery tickets hypothesis.}
The lottery tickets hypothesis of Frankle and Carbin~\cite{frankle2018lottery} explores the possibility of pruning early in training by revealing that some sparse subnetworks inside neural networks can reach accuracy matching that of the full network when trained in isolation.
Setting the weights of the sparse architecture appropriately is critical to reach good performance.
Frankle and Carbin~\cite{frankle2018lottery} provide a proof of concept of the lottery ticket hypothesis on small vision benchmarks while Frankle~\etal~\cite{frankle2019lottery} conduct further experiments with deeper networks, which result in the introduction of rewinding and a consequent revision to the original hypothesis.
Rewinding indeed consists in resetting the parameters to their value ``early in training'' rather than their value from before training.
Liu~\etal~\cite{liu2018rethinking} question the importance of weights resetting and observe for moderate pruning rates and without rewinding that the pruned architecture alone is responsible for successful training.
Zhou~\etal~\cite{zhou2019deconstructing} conduct ablation studies on the lottery tickets hypothesis and show, among others, the importance of the signs of the reset weights.
Yu~\etal~\cite{yu2019playing} investigate the lottery ticket hypothesis in reinforcement learning problems.
These works aim at better understanding winning ticket initializations properties; however none of them investigate their relationship with self-supervised tasks.
Interestingly, Morcos~\etal~\cite{morcos2019one} show that winning tickets initializations transfer across different image classification datasets, thus suggesting that winning tickets do not entirely overfit to the particular data distribution on which they are found. 
\\

\noindent\textbf{Learning without supervision.}
In self-supervised learning, a network is trained on a pretext task that does not require any manual annotations.
Two main broad types of self-supervised learning approaches appear in the litterature.
The first one consists of methods where the pretext task is created by manipulating the input data.
This includes predicting relative spatial location, colorizing grayscale images or predicting the rotation applied to an image~\cite{doersch2015unsupervised,wang2015unsupervised,zhang2016colorful,noroozi2016unsupervised,pathak2017learning,gidaris2018unsupervised,tian2019contrastive,henaff2019data}.
The second one is composed of methods~\cite{dosovitskiy2016discriminative,bojanowski2017unsupervised,wu2018unsupervised,caron2018deep} where images are treated as different instances that should be discriminated from one another.
Representations learnt using self-supervision are most often evaluated via transfer learning to a supervised task.
The better the pre-training with self-supervised learning, the better the performance on the transfer task.
In this work, we broaden the scope of evaluation of self-supervised methods by evaluating subnetworks pruned with these pretext tasks.

%% file: approach.tex
\begin{figure*}[t!]
\centering
\includegraphics[width=0.33\linewidth]{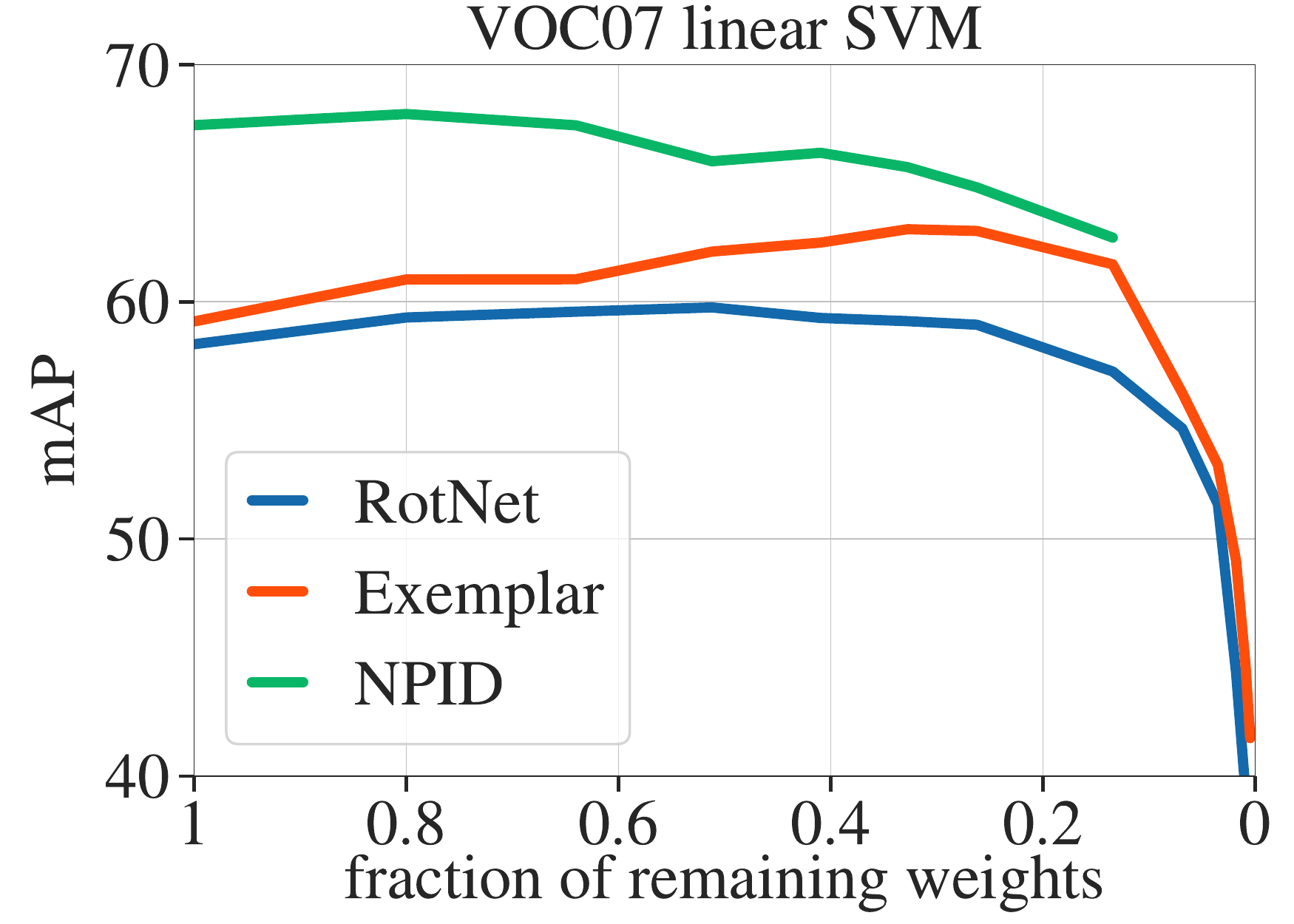}
\includegraphics[width=0.33\linewidth]{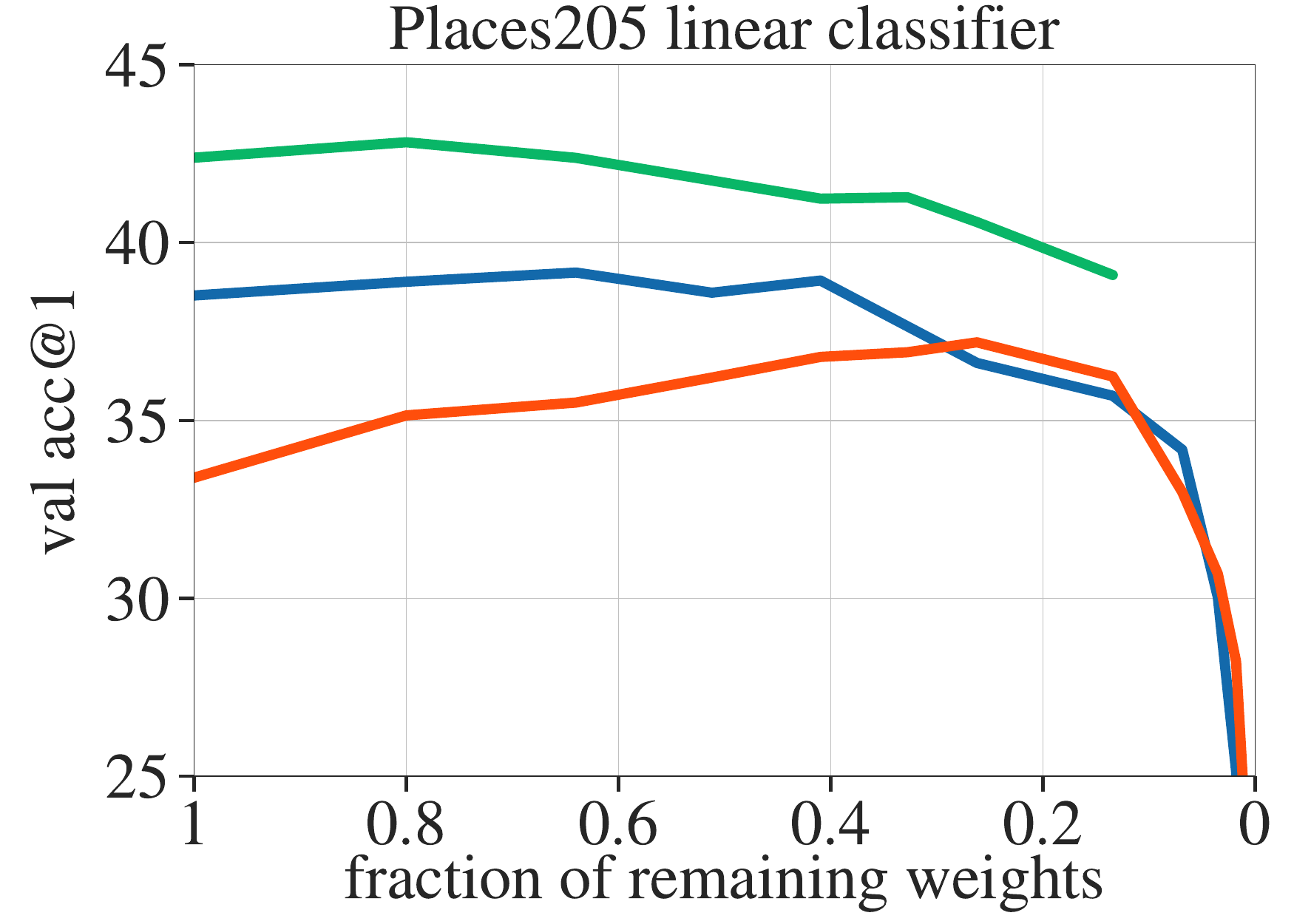}
\includegraphics[width=0.33\linewidth]{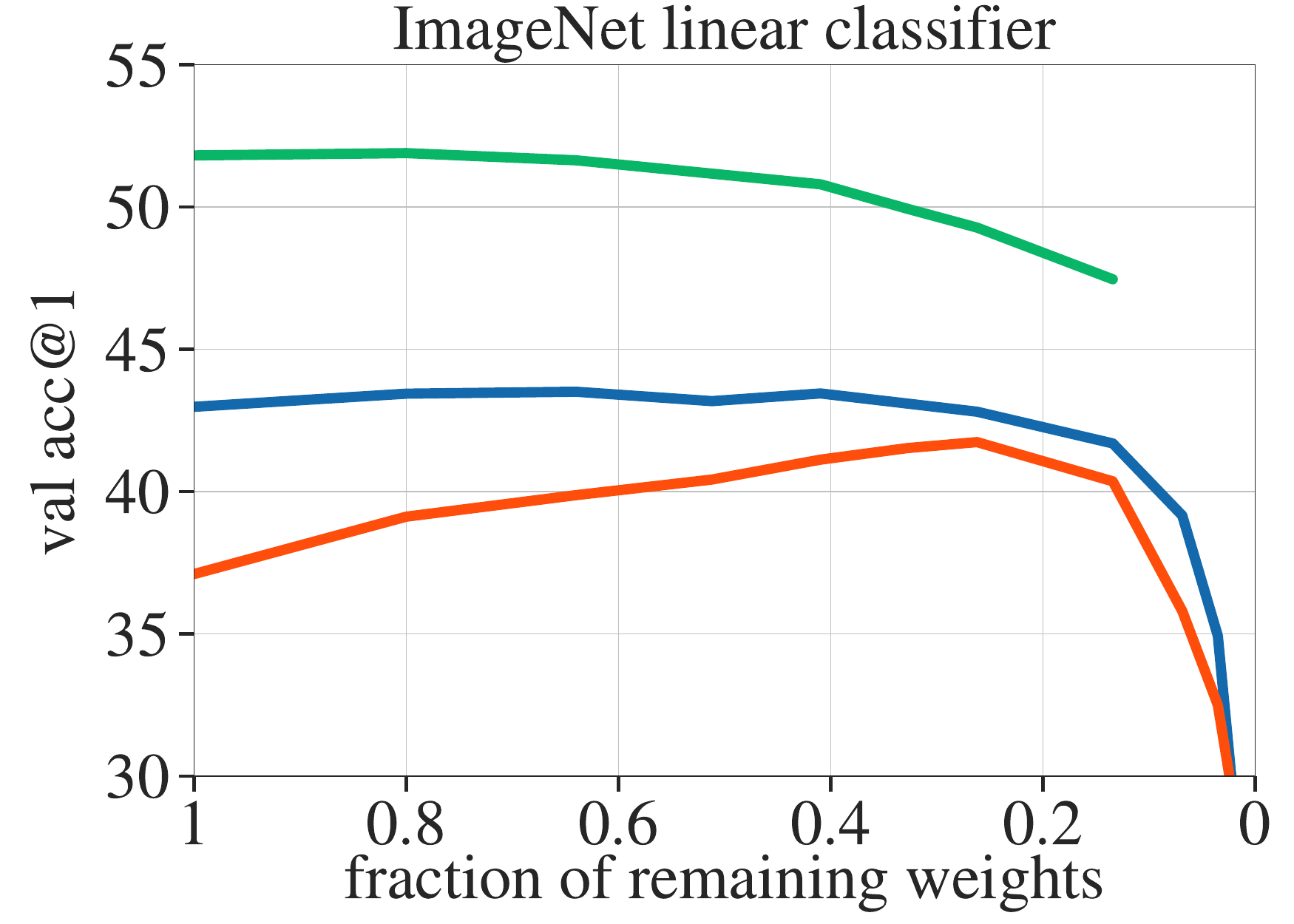}
\caption{
Transfer learning performance of pruned representations from self-supervised methods as we vary the pruning rate.
We show performance when training linear SVMs on VOC2007 and linear classifiers with mini-batch stochastic gradient descent on Places205 and ImageNet.
For reference, on VOC$07$ dataset, the performance for supervised ImageNet features and for random features are respectively $88$mAP and $7.7$mAP~(numbers from Goyal~\etal~\cite{goyal2019scaling}).
We do not prune NPID networks for extreme rates (i.e. less than $0.1$ of remaining weights) because this is too computationally intensive.
}
\label{fig:eval}
\end{figure*}
\section{Approach}

In this work, our goal is to study how a standard pruning method primarily developed for supervised learning applies to networks trained without annotations.
We use well-established methods from the unstructured pruning and self-supervised literatures to do so.
Specifically, we adopt the magnitude-based unstructured iterative pruning process of Han~\etal~\cite{han2015learning} to extract sparse subnetworks from an over-parameterized network.
Following recent works, we reset the subsequent subnetworks to a selected set of weights~\cite{frankle2018lottery,frankle2019lottery} or randomly re-initialize them~\cite{liu2018rethinking}.
The self-supervised tasks we consider are rotation classification of Gidaris~\etal~\cite{gidaris2018unsupervised} and the ``Exemplar'' approach of Dosovitskiy~\etal~\cite{dosovitskiy2016discriminative}.
We provide details about our implementation at the end of this section.
\\

\noindent\textbf{Preliminaries.}
We represent a subnetwork $(W, m)$ by the association of a mask $m$ in $\{0, 1\}^d$ and weight $W$ in $\mathbb{R}^{d}$.
The convolutional network, or convnet, function associated with a subnetwork is denoted by $f_{m \odot W}$, where $\odot$ is element-wise product.
We refer to the vector obtained at the penultimate layer of a convnet as feature or representation.
Such a representation is pruned if $m$ has at least one zero component.
The subnetwork weights $W$ may be \emph{pre-trained}, such that the corresponding feature function can be transferred to downstream tasks.
On the other hand, the subnetwork weights $W$ may be \emph{initialization weights}, such that the corresponding convnet $f_{m \odot W}$ can be re-trained from scratch.

\subsection{Unstructured magnitude-based pruning}

\noindent\textbf{Pruned mask.}
Han~\etal~\cite{han2015learning} propose an algorithm to prune networks by estimating which weights are important.
This approach consists of compressing networks by alternatively minimizing a training objective and pruning the network parameters with the smallest magnitude, hence progressively reducing the network size.
At each pruning iteration, the network is first trained to convergence, thus arriving at weights $W^*$.
Then, the mask $m$ is updated by setting to zero the elements already masked plus the smallest elements of $\{ |W^*[j]|\, |\,m[j] \neq 0 \}$.
\\

\noindent\textbf{Weight resetting.}
Frankle and Carbin~\cite{frankle2018lottery} refine this approach and propose to also find a good initialization $W$ for each subnetwork such that it may be re-trained from scratch.
On small-scale computer vision datasets and with shallow architectures, they indeed show that sub-architectures found with iterative magnitude pruning can be re-trained from the start, as long as their weights are reset to their initial values.
Further experiments, however, have shown that this observation does not exactly hold for more challenging benchmarks such as ImageNet~\cite{frankle2019lottery,liu2018rethinking}.
Specifically, Frankle~\etal~\cite{frankle2019lottery} found that resetting weights to their value from an early stage in optimization can still lead to good trainable subnetworks.
Formally, at each pruning iteration, the subnetwork is reset to weights $W_k$ obtained after $k$ weight updates from the first pruning iteration.
Liu~\etal~\cite{liu2018rethinking}, on the other hand, argue that the mask $m$ only is responsible for the good performance of the subnetwork and thus its weights $W$ may be randomly drawn at initialization.
In our work, we consider both weights initialization schemes: winning tickets of Frankle~\etal~\cite{frankle2019lottery} or random re-initialization~\cite{liu2018rethinking}.
\\

\subsection{Self-supervised learning}
We prune networks without supervision by simply setting the training objective in the method of Han~\etal~\cite{han2015learning} to a self-supervised pretext task.
We consider two prominent self-supervised methods: RotNet \cite{gidaris2018unsupervised} and the Exemplar approach of Dosovitskiy~\etal~\cite{dosovitskiy2016discriminative} following the implementation of Doersch~\etal~\cite{doersch2017multi}.
RotNet consists in predicting the rotation which was applied to the input image among a set of $4$ possible large rotations: $\{0\degree, 90\degree, 180\degree, 270\degree\}$.
Exemplar is a classification problem where each image and its transformations form a class, leading to as many classes as there are training examples.
We choose these two self-supervised tasks because they have opposite characteristics:
RotNet encourages discriminative features to data transformations and has a small number of classes, while Exemplar encourages invariance to data transformations and its output space dimension is large.
We also investigate the non-parametric instance discrimination (NPID) approach of Wu~\etal~\cite{wu2018unsupervised}, which is a variant of the Exemplar method that uses a non-parametric softmax layer and a memory bank of feature vectors.

\begin{figure*}[t!]
\centering
\includegraphics[width=0.33\linewidth]{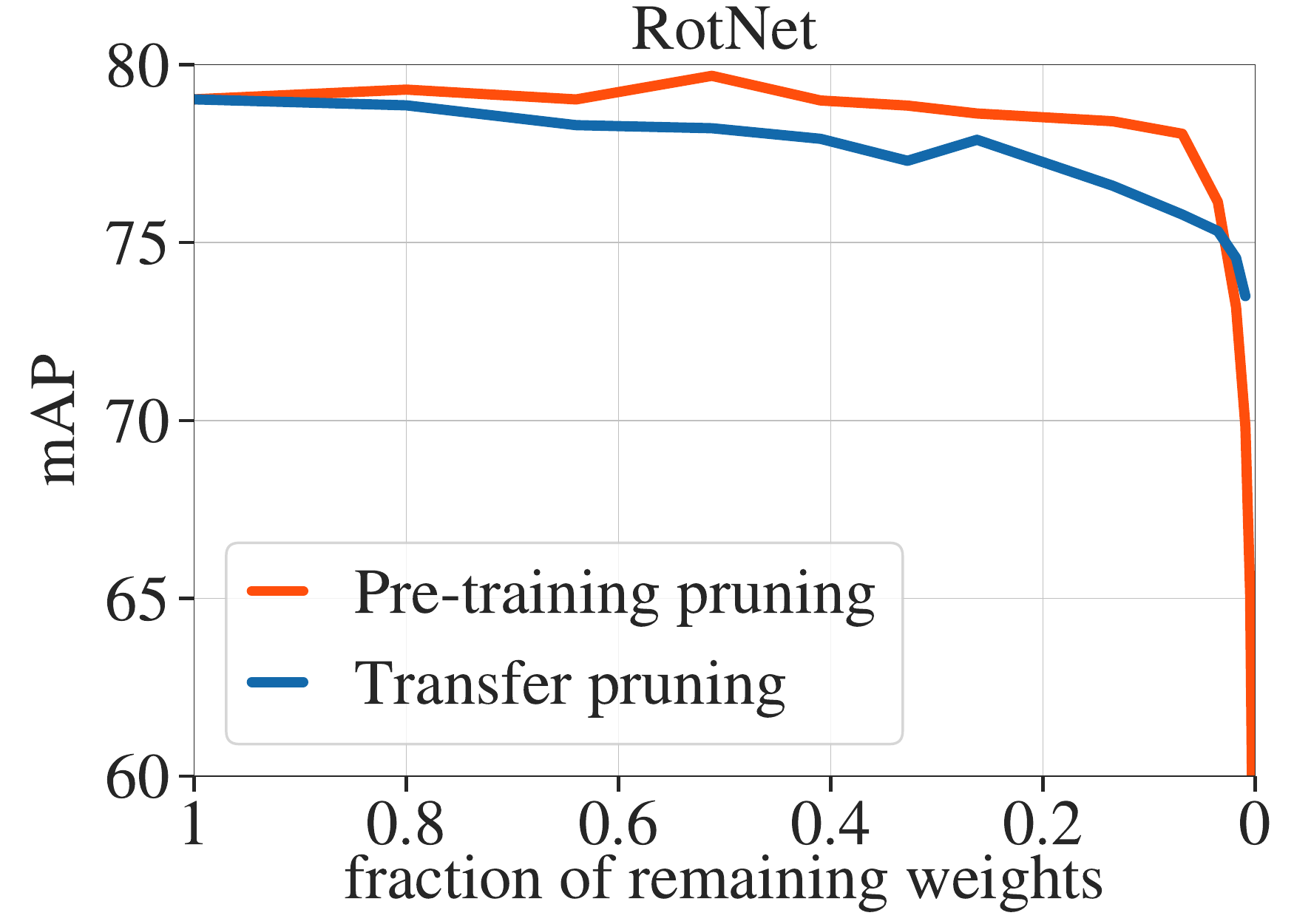}
\includegraphics[width=0.33\linewidth]{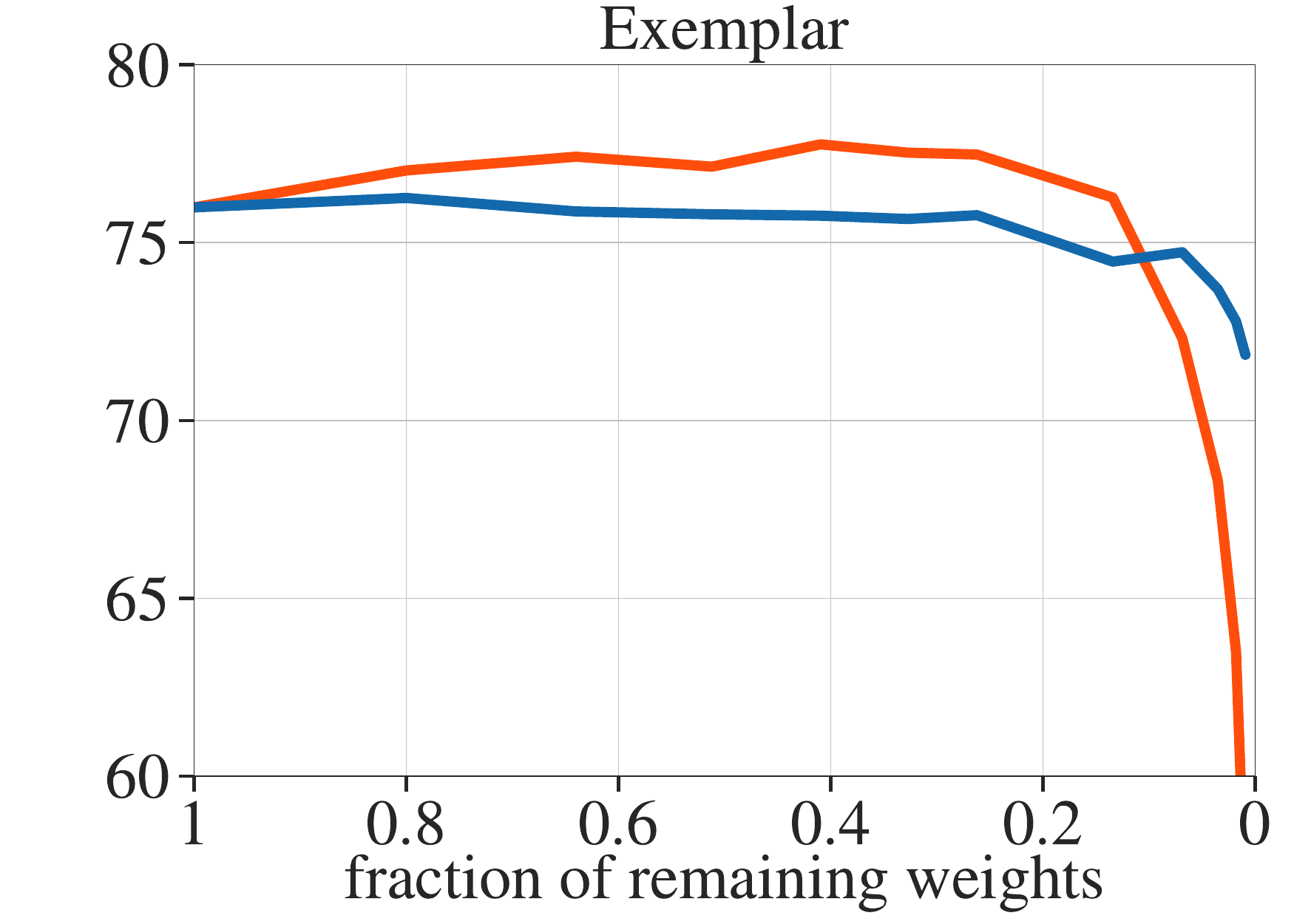}
\includegraphics[width=0.33\linewidth]{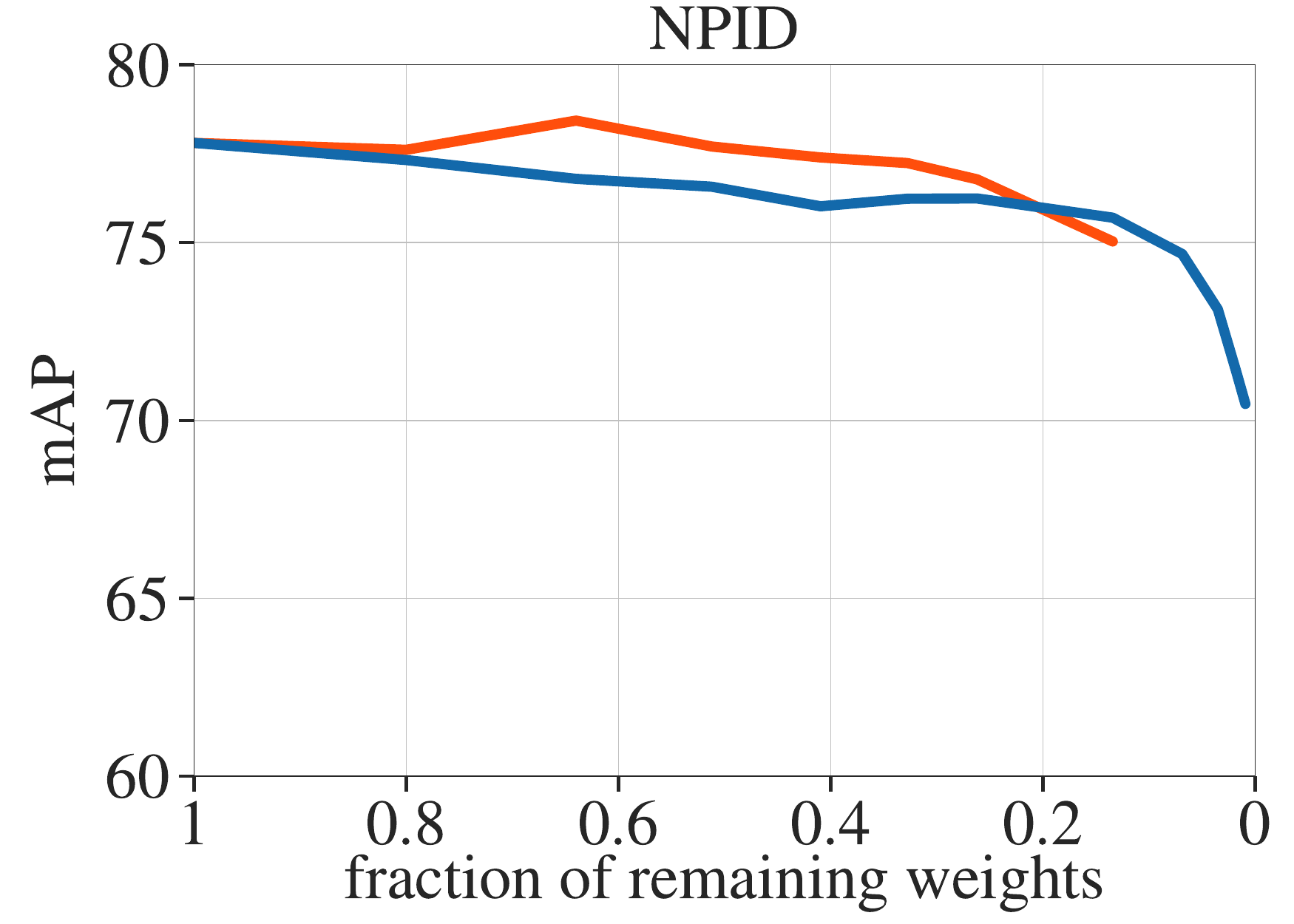}
\caption{
Transfer learning performance of representations pruned during pre-training or during transfer on VOC07 classification task with full fine-tuning as we vary the pruning rate.
For reference, finetuning unpruned supervised ImageNet features gives $90.3$ while training from random initialization gives $48.4$~(numbers from Goyal~\etal~\cite{goyal2019scaling}).
}
\label{fig:pretrain_or_transfer}
\end{figure*}

\subsection{Implementation}
\noindent\textbf{Pruning.}
We follow closely the winning tickets setup of Morcos~\etal~\cite{morcos2019one}.
At each pruning iteration, we globally prune $20\%$ of the remaining weights.
The last fully-connected and batch-norm layers parameters are left unpruned.
We apply up to $30$ pruning iterations to reach extreme pruning rates where only $0.1\%$ of the weights remain.
Overall we report results for $14$ different pruning rates ranging from $20\%$ to $99.9\%$, thus covering both moderate and extreme sparsity.
The weight resetting parameter is set to $3 \times 1,3$M samples, which corresponds to $3$ epochs on full ImageNet.
More details about this late resetting parameter are in the supplementary material.
\\

\noindent\textbf{Datasets and models.}
In this work, we choose to mostly work with ImageNet dataset~\cite{deng2009imagenet}, though we also report some results for CIFAR-10~\cite{krizhevsky2009learning}.
We use ResNet-50 on ImageNet and ResNet-$18$ for CIFAR-10~\cite{he2016deep}.
In supplementary material, we also report results with more architectures: AlexNet~\cite{krizhevsky2012imagenet} and the modified VGG-19~\cite{simonyan2014very} of Morcos~\etal~\cite{morcos2019one} (multilayer perceptron (MLP) is replaced by a fully connected layer).
Our experiments on ImageNet are computationally-demanding since pruning can involve training deep networks from scratch up to $31$ times.
For this reason, we distribute most of our runs across several GPUs.
Models are trained with weight decay and stochastic gradient descent with a momentum of $0.9$.
We use PyTorch~\cite{paszke2017automatic} version 1.0 for all our experiments.
Full training details for each of our experiments are in the supplementary material.
We run each experiment with $6$ (CIFAR-$10$) or $3$ (ImageNet) random seeds, and show the mean and standard error of the accuracy.

%% file: experiments.tex
\section{Experimental study}

In our experimental study, we first evaluate the pruned representations: we investigate the effect of pruning on the subsequent self-supervised representations.
We also evaluate pruning depending on whether it is performed with or without supervision.
Finally, we explore the impact of adding a little supervision during pruning.

\subsection{Evaluating Pruned Self-Supervised Features} \label{sec:eval_features}
In this section, we evaluate the quality of pruned self-supervised features by transferring them to different downstream tasks.
We show that pruning self-supervised features preserves their transfer performance for a wide range of pruning rates.
\\

\noindent\textbf{Feature transfer performance when pruning.}
In this experiment, we are interested in the features transferability as we vary the amount of pruned weights in the representation function.
We follow part of the benchmark proposed by Goyal~\etal~\cite{goyal2019scaling} for evaluating unsupervised representations.
In particular, we focus on training linear classifiers for VOC$07$~\cite{everingham2010pascal}, Places$205$~\cite{zhou2014learning} and ImageNet~\cite{deng2009imagenet} labels classification tasks on top of the final representations given by a pre-trained pruned convnet.
In Figure~\ref{fig:eval}, we observe that pruning up to $90\%$ of the networks weights (i.e. there is less than $0.1$ remaining weights) does not deteriorate the resulting features quality when evaluated with linear classifier on VOC$07$, Places$205$ and ImageNet datasets.
Surprisingly, we observe that for RotNet and Exemplar self-supervised approaches, pruning the features even improves slightly their transfer performance.
An explanation may be that by pruning, task-specific information is removed from the features which leads to a better transferability.
However, this is not the case for the best performing self-supervised method NPID: the quality of the representation remains constant when pruning at moderate rates.
In any case, when features are severely pruned (i.e. less than $5\%$ of the weights remain), their quality drops significantly.
\\

\noindent\textbf{Pruning during pre-training or during transfer?}
One difficulty of pruning while pre-training is that the training objective used to prune the network (i.e. the pre-training task) is not directly related to the transfer task.
Thus, we study in this experiment if pruning directly using the target objective might be a better strategy to obtain good performance on this same target task.
In Figure~\ref{fig:pretrain_or_transfer}, we evaluate pruned unsupervised features by transferring them to Pascal VOC 2007~\cite{everingham2010pascal} classification task.
The pre-trained features are used as weight initialization, then all the network parameters are trained (i.e. finetuned) on VOC$07$.
The features are either pruned during pre-training (i.e. on RotNet or Exemplar self-supervised tasks) or during transfer (i.e. when finetuning on Pascal VOC$07$ classification task directly).
On Pascal VOC$07$, we train the models for $90$ epochs on combined train and val sets, starting with a learning rate of $0.01$ decayed by a factor $10$ at epochs $50$, $65$ and $80$; we report test central crop evaluation.
We observe in Figure~\ref{fig:pretrain_or_transfer} that finetuning a pruned pre-trained network gives better transfer performance than finetuning while pruning.
This observation has an interesting real-world impact: pruned pre-trained models can be released for users with limited computational budget without the need of additional pruning on the transfer task from their side.
\\

\begin{figure*}[t]
\centering
\subfloat[ImageNet]{
\includegraphics[width=0.24\linewidth]{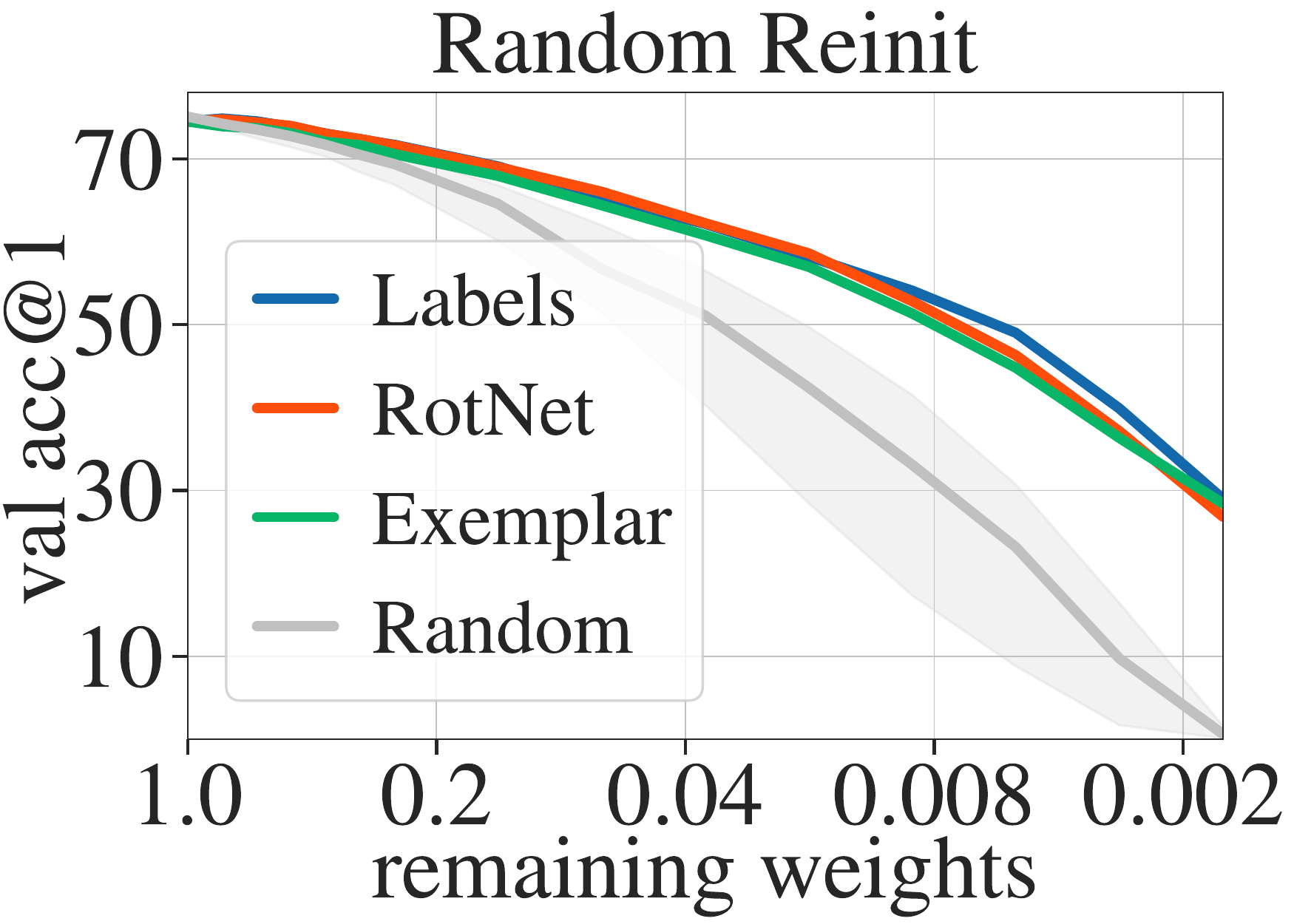}
\includegraphics[width=0.24\linewidth]{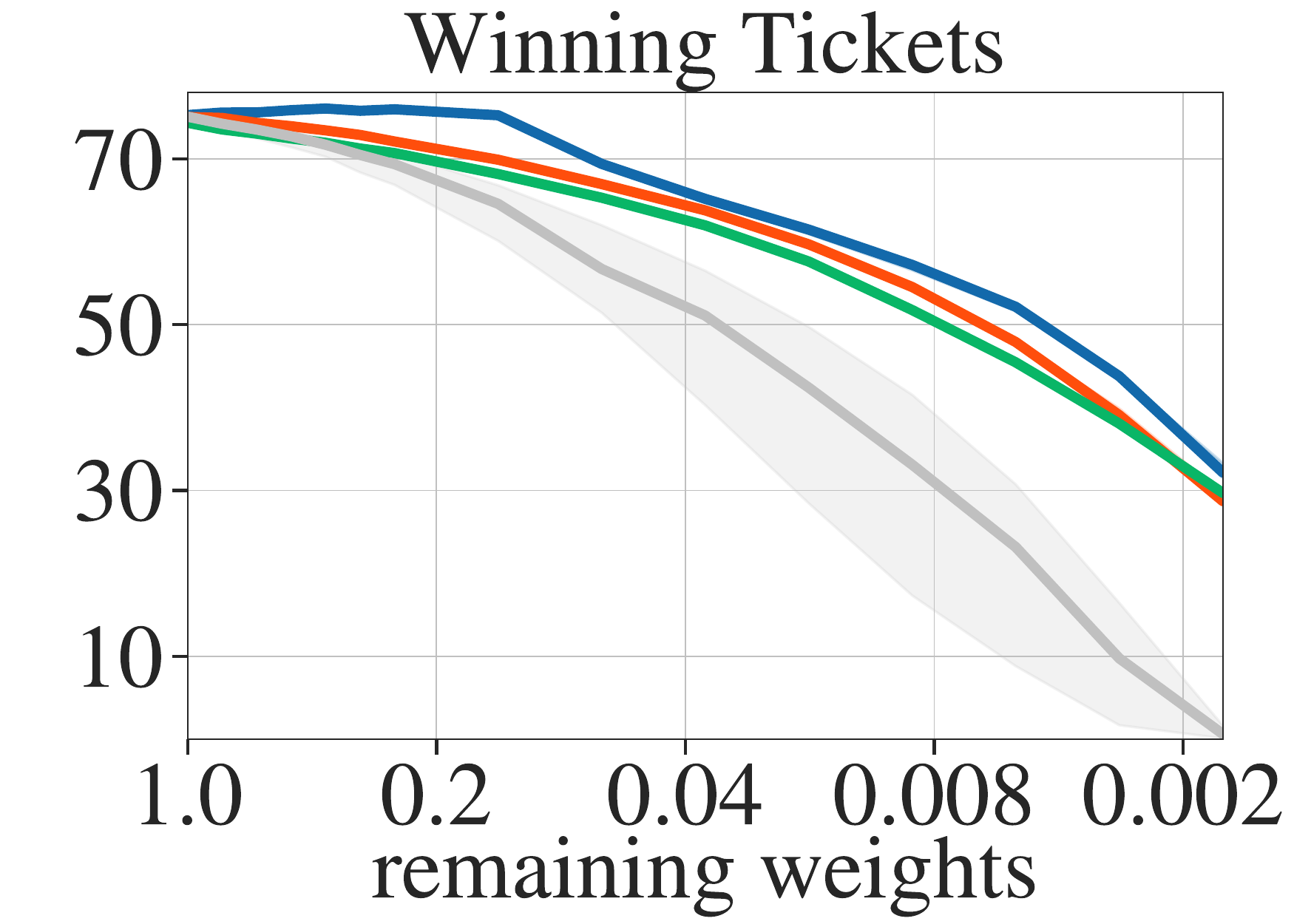}
\label{fig:imnet}
}
\subfloat[CIFAR-10]{
\includegraphics[width=0.24\linewidth]{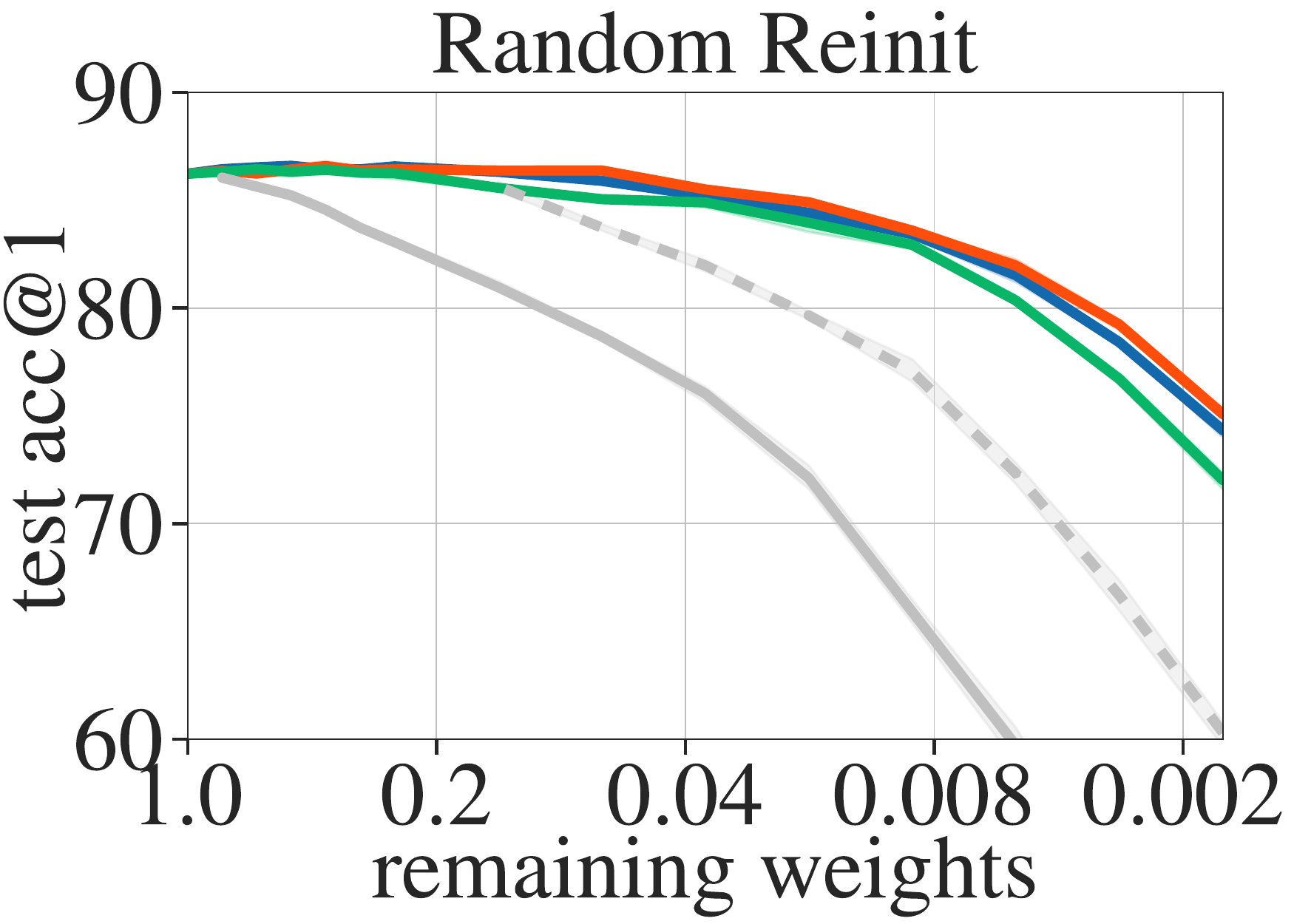}
\includegraphics[width=0.24\linewidth]{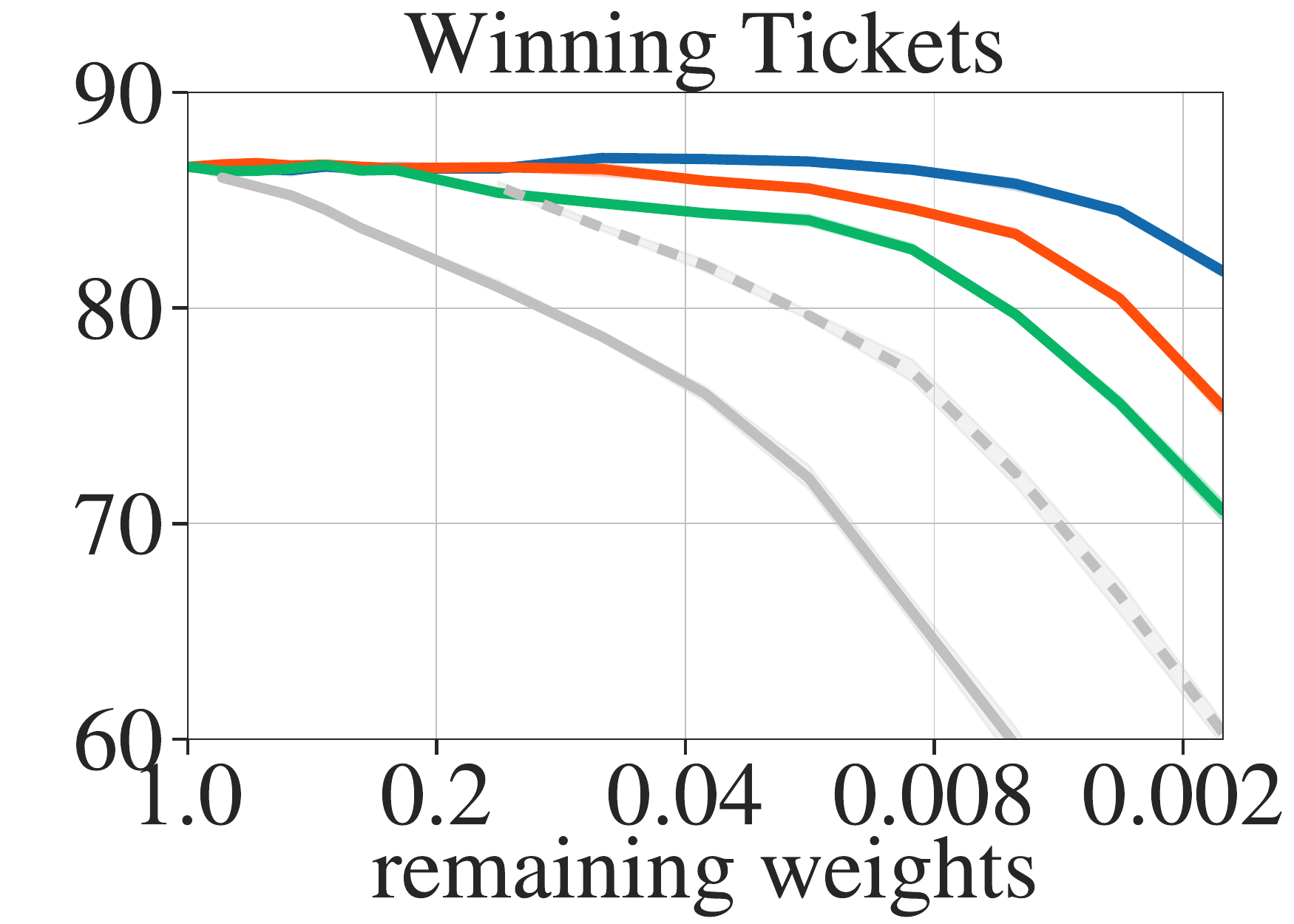}
\label{fig:cifar10}
}
\caption{
We report ImageNet val~\protect\subref{fig:imnet} and CIFAR-10 test~\protect\subref{fig:cifar10} top-$1$ accuracy when re-training subnetworks pruned with self-supervised tasks: RotNet or Exemplar.
The x-axis corresponds to different rates of remaining weights.
We use two different schemes for weight initialization:
``Random Reinit'': random re-initialization of the subnetwork~\cite{liu2018rethinking};
``Winning Tickets'': inherited from early phase of training following the lottery tickets hypothesis~\cite{frankle2019lottery}.
We also show performance when re-training subnetworks pruned with labels (``Labels'') or randomly pruned subnetworks (``Random'') which consist of randomly permuted masks and randomly drawn weights from the initialization distribution.
On CIFAR-$10$, deep models are highly sparse with only approximately $\sim15\%$ of non-zero weights.
Thus, we adjust the random baseline (dashed curve) to start with the correct mask at the natural level of network sparsity (details in Section~\ref{sec:caveat}).
}
\label{fig:randomreinit_wt}
\end{figure*}
\subsection{Evaluating Self-Supervised Masks}
In the previous section we evaluate pruned \emph{representations}.
In this section, we evaluate more specifically the \emph{masks} and corresponding \emph{weight initializations} obtained from self-supervised pruning.
In particular, we show that these can be re-trained successfully on a supervised task on the same dataset.
Finally, we investigate the effect of adding some label supervision during pruning.

\subsubsection{Evaluating Self-Supervised Pruning} \label{sec:nolabel}
In this set of experiments, we evaluate the quality of the pruned masks and corresponding weight initializations obtained by pruning with a self-supervised objective.
\\

\noindent\textbf{Experimental setting.}
We evaluate the pruned subnetworks by training them from their initialization on semantic label classification.
The difficulty is the following: unlike the subnetworks pruned with labels, the ones pruned with self-supervised tasks have not seen the labels during pruning and are consequently fully ``label-agnostic''.
We compare them to networks pruned with labels directly and to randomly pruned subnetworks re-trained on label classification.
For the remaining of the paper, we use a logarithmic scale for the pruning rates axis in order to focus on situations where very few weights remain.
Indeed, for moderate pruning rates, a random pruning baseline is competitive with supervised pruning.
Moreover, we observe that pruning deep networks on CIFAR-$10$ is trivial for moderate rates: we detail this finding in Section~\ref{sec:caveat}.
\\

\noindent\textbf{Re-training self-supervised masks on label classification.}
In Figure~\ref{fig:randomreinit_wt}, we show validation accuracy of pruned subnetworks re-trained on label classification, at different pruning ratios.
We show results with ResNet-$50$ for ImageNet and ResNet-$18$ for CIFAR-$10$ and investigate more architectures in the supplementary material with similar conclusions.
We observe in Figure~\ref{fig:randomreinit_wt} that subnetworks pruned without any supervision are still capable of reaching good accuracy when re-trained on label classification.
When the subnetworks start from random initialization (``Random Reinit''), self-supervised pruning even matches the performance of supervised pruning.
However, with winning tickets initializations (``Winning Tickets''), subnetworks pruned without labels perform significantly below supervised pruning, but remain way above the random pruning level.
Note that the gap of performance between self-supervised pruning and random pruning increases when the network is severely pruned.
Overall, this experiment suggests that self-supervised pruning gives pruned masks and weight initializations which are much better than random pruning when evaluated on label classification, and even on par with labels pruning when randomly re-initialized.
\\

\noindent\textbf{Winning tickets versus random initialization.}
For pruning with supervision (``Labels'' curves in Figure~\ref{fig:randomreinit_wt}), we observe that our results provide further empirical evidence to the lottery tickets hypothesis of Frankle~\etal~\cite{frankle2019lottery}.
Indeed, we observe in Figure~\ref{fig:randomreinit_wt} that resetting the weights (winning tickets strategy) of the masks pruned with labels gives significantly better accuracy compared to random re-initialization.
Interestingly, this is not the case for subnetworks pruned without labels:
winning tickets initialization gives only a very slight boost (or even no boost at all) of performance compared to randomly re-initialization for subnetworks pruned with RotNet or Exemplar self-supervised tasks.
Overall, these results suggest that the quality of the pruned mask $m$ itself is similar for supervised or self-supervised pruning but the \textit{weights} of self-supervised subnetworks are not as good starting points as the ones inherited from label classification task directly.
\\

\begin{figure*}[t]
\centering
  \begin{tabular}{cccc}
\includegraphics[width=0.22\linewidth]{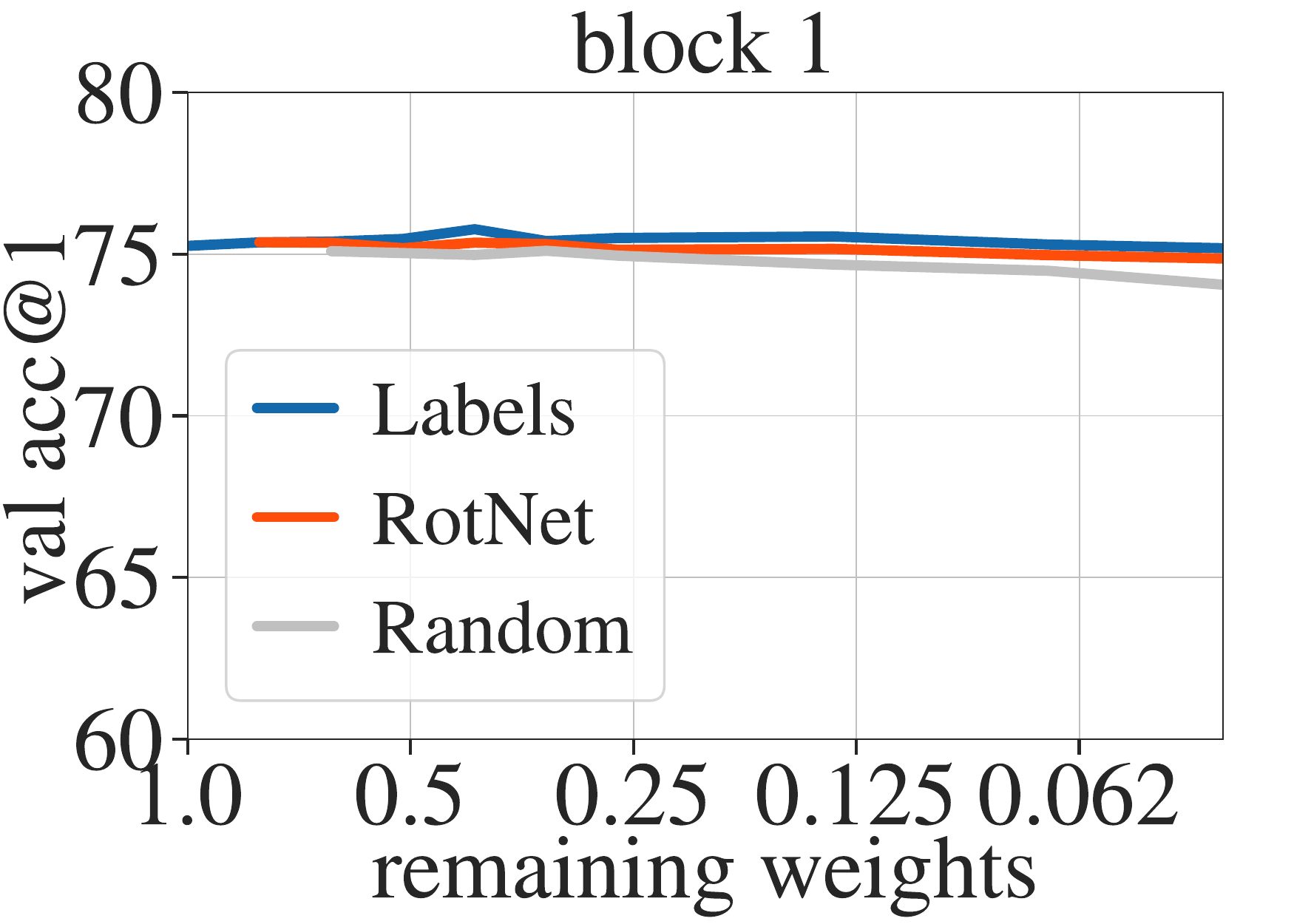} &
\includegraphics[width=0.22\linewidth]{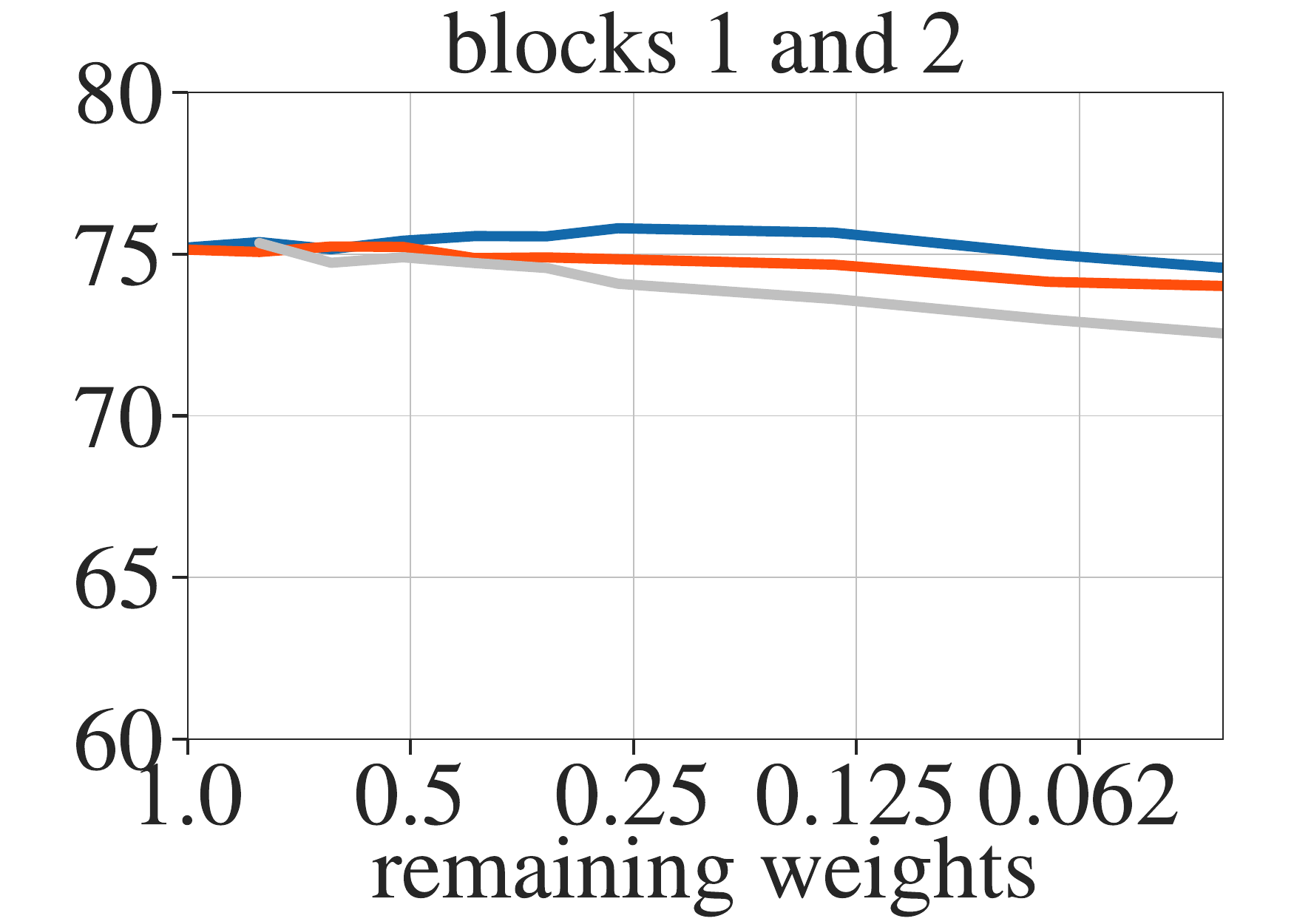} &
\includegraphics[width=0.22\linewidth]{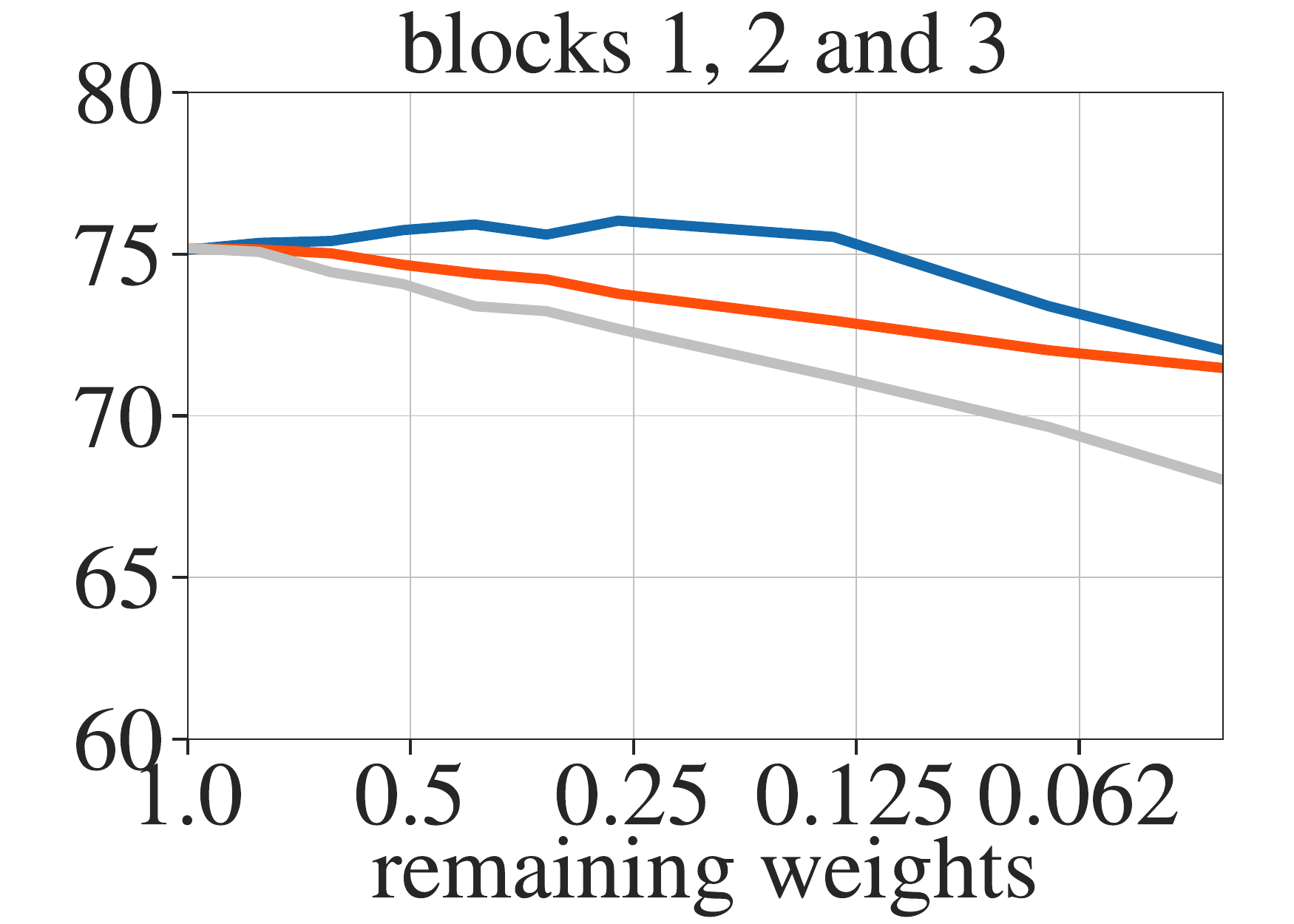} &
\includegraphics[width=0.22\linewidth]{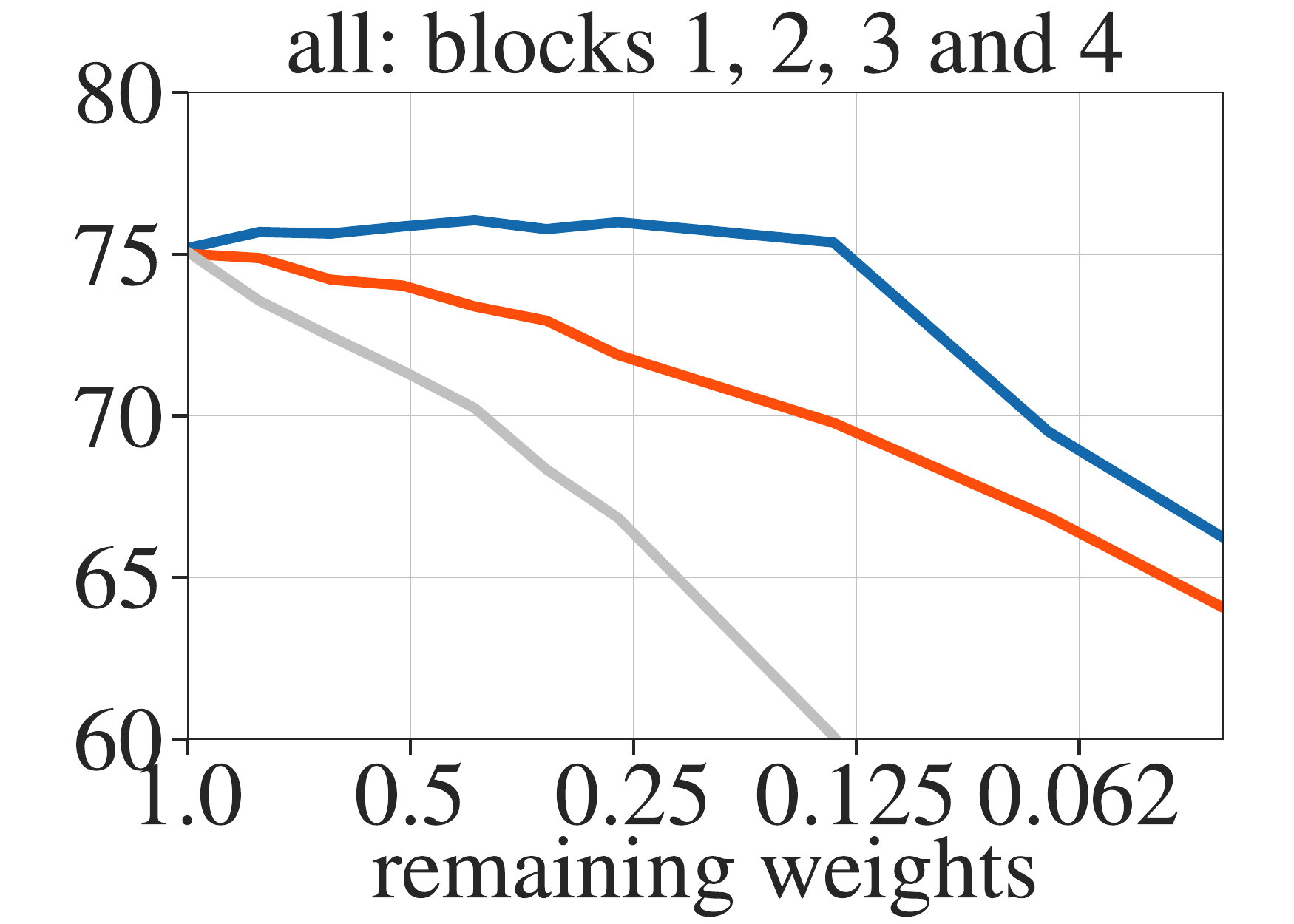}
  \end{tabular}
\caption{
ImageNet top $1$ validation accuracy of winning tickets found by pruning partly or entirely (all) a network with labels classification or RotNet.
We show for reference results when the corresponding network layers are randomly pruned.
}
\label{fig:layers}
\end{figure*}

\noindent\textbf{Layerwise Pruning.}
We further investigate the difference between supervised and self-supervised winning tickets on ImageNet by looking at their performance as we prune up to a given depth.
In Figure~\ref{fig:layers}, we verify that when pruning only the first convolutional layers this gap remains narrow, while it becomes much wider when pruning higher level layers.
Indeed, it has been observed that most self-supervised approaches produce good shallow and mid-level features but poorer high level features~\cite{jing2019self}.
This means that we should expect the quality of self-supervised pruning to depend on the depth of the pruning.
We confirm this intuition and show the difference between self-supervised and supervised pruning increases with pruning depth.

\subsubsection{Semi-Supervised Pruning} \label{sec:semisup}
Finally in this experiment, we investigate the impact of using part of the labels for pruning.
We find in previous experiments (Section~\ref{sec:nolabel}) that pruned masks uncovered by supervision or self-supervision lead to similar performance when re-trained on labels classification, however, winning tickets initializations from labels are globally better than ones from pretext self-supervised tasks.
We show in this section that adding a little bit of label supervision for pruning improves the resulting winning tickets initializations performance on label classification, suggesting that these initializations are task-dependant to some extent.
\\

\noindent\textbf{Experimental setting.}
In this experiment, we consider $10\%$ of ImageNet labels.
We either sample this labeled subset per class (i.e. we preserve all the classes) or we sample a subset of classes and keep all the images for these classes (i.e. we only have $100$ classes).
Either way, we get approximately $130$k labeled images and use the remaining images without their label.
We use $S^4L$, the semi-supervised setting of Zhai~\etal~\cite{zhai2019s} since they show better performance on ImageNet classification compared to virtual adversarial training~\cite{miyato2018virtual} or pseudo-labeling~\cite{lee2013pseudo}.
In particular, we use the RotNet variant of $S^4L$: the training loss corresponds to the sum of a label classification loss applied to labeled examples only, and a RotNet loss applied to all data samples.
We compare this semi-supervised pruning setting to pruning without any labels (masks from Section~\ref{sec:nolabel}) and to pruning with few labels only without adding a RotNet loss.
We also show for reference the performance of subnetworks obtained with fully-supervised or random pruning.
We follow the same methodology from Section~\ref{sec:nolabel} to evaluate pruning masks: we re-train the subsequent pruned masks and corresponding winning tickets weight initializations on ImageNet classification task.
\\

\begin{figure}[h]
\centering
\subfloat[Preserved \# of classes]{
\includegraphics[width=0.48\linewidth]{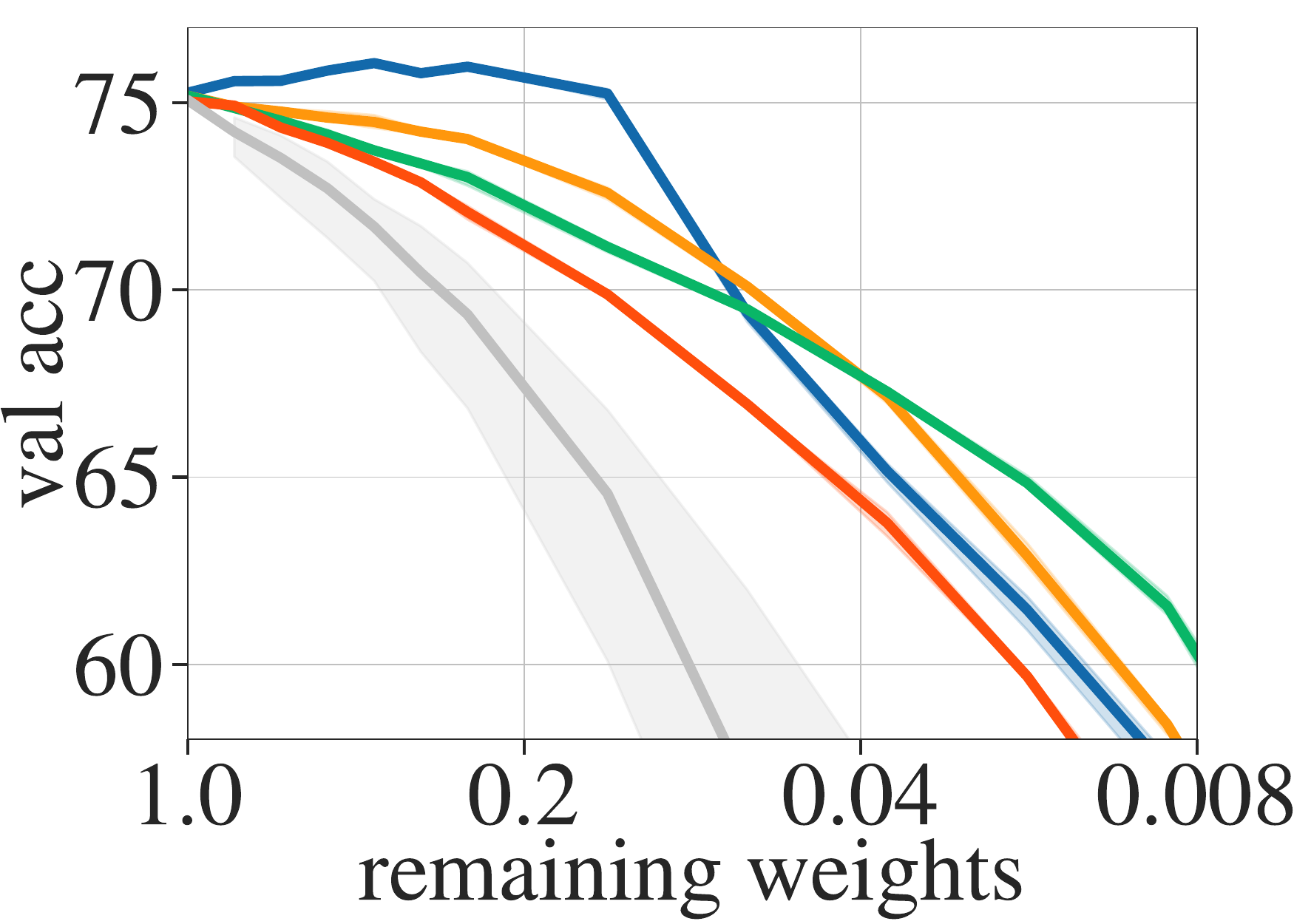}
\label{fig:perclass}
}
\subfloat[Preserved \# of images per class]{
\includegraphics[width=0.48\linewidth]{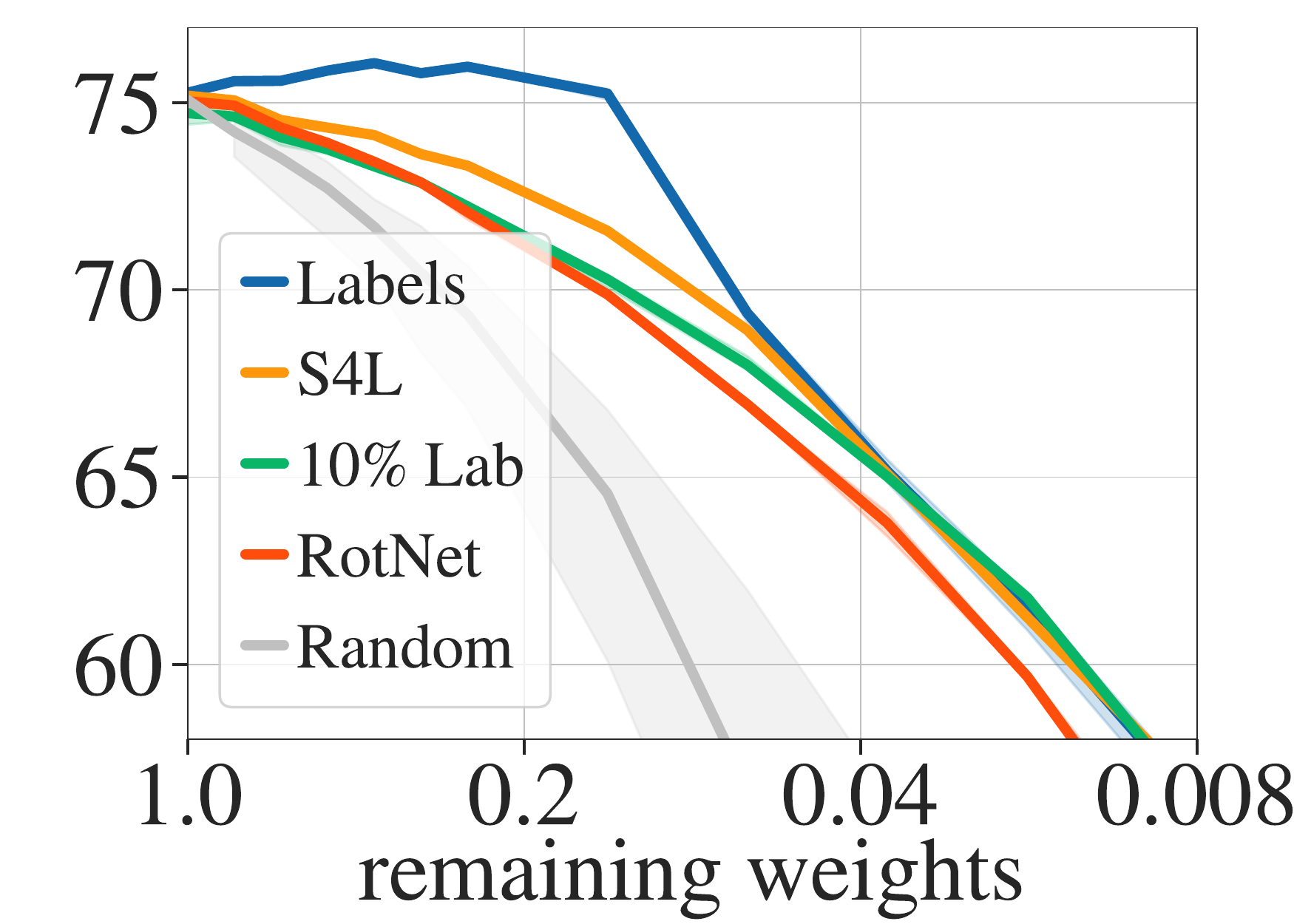}
\label{fig:classes}
}
\caption{
We report ImageNet validation top-$1$ accuracy when re-training subnetworks pruned with varying level of access to labels supervision.
``Labels'' corresponds to classical supervised pruning on all the labels, ``$S^4L$'' is semi-supervised pruning with S4L RotNet method, ``$10\%$ Lab'' is supervised pruning on a reduced dataset and ``RotNet'' corresponds to unsupervised pruning.
We either sample the $10\%$ labeled subset by selecting $10\%$ images per class~\protect\subref{fig:perclass} or by selecting $10\%$ of classes~\protect\subref{fig:classes}.
}
\label{fig:semisup}
\end{figure}
\noindent\textbf{Pruning with limited supervision.}
In Figure~\ref{fig:semisup}, we observe that adding $10\%$ of labels ($S^4L$) for pruning leads to significantly improved performance for the resulting subnetworks compared to pruning with no labels (RotNet).
We previously observed that the pruned masks are of similar quality between supervised and unsupervised pruning (see Figure~\ref{fig:randomreinit_wt}).
Hence, the improvement brought by adding labels is likely to be mainly due to better winning tickets initializations.
Since both semi-supervised pruning and even supervised pruning with few data perform better than pruning with no labels, we conclude that winning tickets initializations are labels-dependant to some extent.
Indeed, we surprisingly observe that simply using $10\%$ labeled images and nothing else for pruning (``$10\%$ Lab'' in Figure~\ref{fig:semisup}) provides better winning tickets initializations than pruning on the entire, unlabeled dataset (``RotNet'').
Interestingly, we observe that this gap of performance is much wider when winning tickets are inherited from a preserved set of classes (Fig.~\ref{fig:perclass}) rather than a reduced one (Fig.~\ref{fig:classes}), which suggests that winning tickets initializations are somewhat ``task''-dependant.
Finally, subnetworks pruned in the semi-supervised setting (``$S^4L$'') perform better in general than both the unsupervised (``RotNet'') and low-data (``$10\%$ Lab'') ones which leads to think that their respective properties capture different statistics that add up to generate better winning tickets initializations.

\subsection{Caveat about pruning deep networks on CIFAR-10} \label{sec:caveat}

\begin{figure}[h]
\centering
\includegraphics[width=0.8\linewidth]{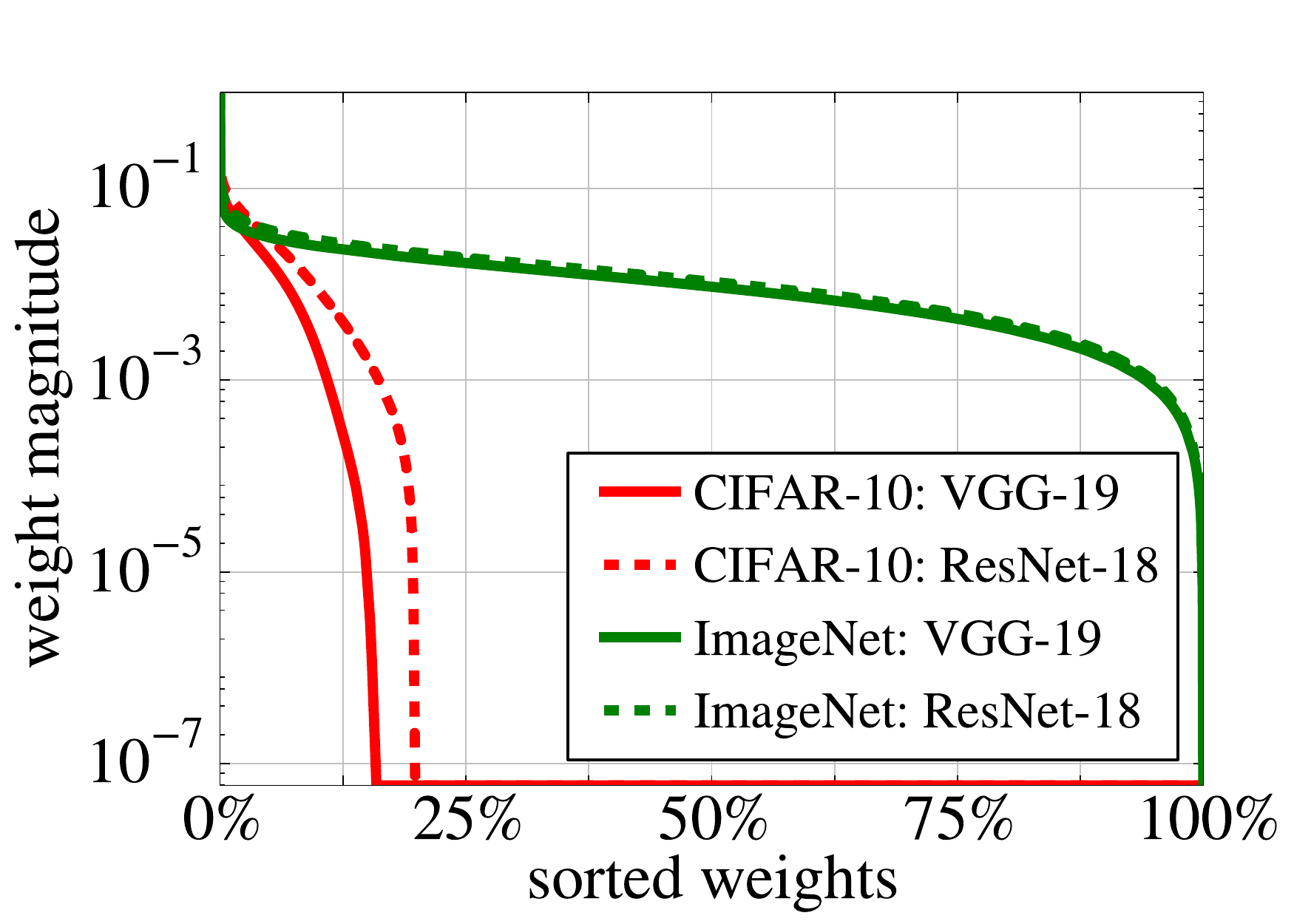}
\caption{
Magnitude of the weights of a trained network on two different datasets: CIFAR-$10$ (green) and ImageNet (red).
We perform thresholding at machine precision value (bottom of y-axis).
On CIFAR-$10$, a trained VGG-$19$ is $84.5\%$ sparse while a trained ResNet-$18$ is $80.3\%$ sparse.
}
\label{fig:ratios}
\end{figure}

Somewhat surprisingly, we find that prior to any pruning, a large proportion of the weights of a deep architecture trained on CIFAR-10 has converged naturally to zero during training.
For example, we observe in Figure~\ref{fig:ratios} that $\sim85\%$ of the weights of a VGG-19 and $\sim80\%$ of that of a ResNet-$18$ at convergence are zeroed (we show results for more architectures in the supplementary material).
As a result, it is trivial to prune these networks without any loss in accuracy.
Unstructured magnitude-based pruning acts here as \textit{training} since are freezed to zero weights that were going to zero anyway~\cite{zhou2019deconstructing}.
Overall, while pruning on CIFAR-10 large networks originally tuned for ImageNet at rates above their natural level of sparsity ($\sim80\%$) is still meaningful, analyzing pruning below this rate may not be conclusive.

In the random global pruning baseline (which can remove non zero weights) of Figure~\ref{fig:randomreinit_wt}, pruning at rates below the natural sparsity of the network degrades accuracy, while pruning of weights that are already zeroed has no effect.
Inconveniently, this performance gap carries over to higher pruning rates (in which we are interested in) and can lead to misleading interpretations.
For fair comparison, we adjust the random mask baseline in Figure~\ref{fig:randomreinit_wt}: we remove this effect by first pruning the weights that naturally converge to zero after training.
Then, we randomly mask the remaining non-zeroed weights to get different final desired pruning rates.
The remaining non-masked weights are randomly initialized.
This baseline therefore corrects for the natural sparsity present in CIFAR-10 networks.

%% file: discussion.tex
\section{Conclusion} \label{sec:discussion}

Our work takes a first step into studying the pruning of networks trained with self-supervised tasks.
We believe this is an emergent and important problem due to the recent rise of highly over-parametrized unsupervised pre-trained networks.
In our study, we empirically provide different insights about pruning self-supervised networks.
Indeed, we show that a well-established pruning method for supervised learning actually works well for self-supervised networks too, in the sense that the quality of the pruned representation is not deteriorated and the pruned masks can be re-trained to good performance on ImageNet labels.
This is somewhat surprising given that labels are not seen during pruning and given that the goal of the pruning algorithm we use is to preserve the performance on the training task, which is agnostic to downstream tasks or ground-truth labels.

We also find several limitations to our study.
First, we have observed pruning through the scope of unstructured magnitude-based pruning only.
Future work might generalize our observations to a wider range of pruning methods, in particular structured pruning.
Second, we have observed while conducting our experiments that winning tickets initializations are particularly sensitive to the late resetting parameter (see the supplementary material for a discussion about our choice of rewind parameter).
The definition of ``early in training'' is somehow ill-defined: network weights change much more for the first epochs than for the last ones.
Thus, by resetting weights early in their optimization, they contain a vast amount of information.
Third, we find that pruning large modern architectures on CIFAR-$10$ should be done with caution as these networks tend to be sparse at convergence, making unstructured pruning at rates below $80\%$ particularly simple.

%% file: appendices.tex
\section{Supplementary Material}
\addcontentsline{toc}{section}{Supplementary Material}
\renewcommand{\thesubsection}{\Alph{subsection}}

\subsection{Late resetting parameter}
\begin{figure}[h]
\centering
\includegraphics[width=0.8\linewidth]{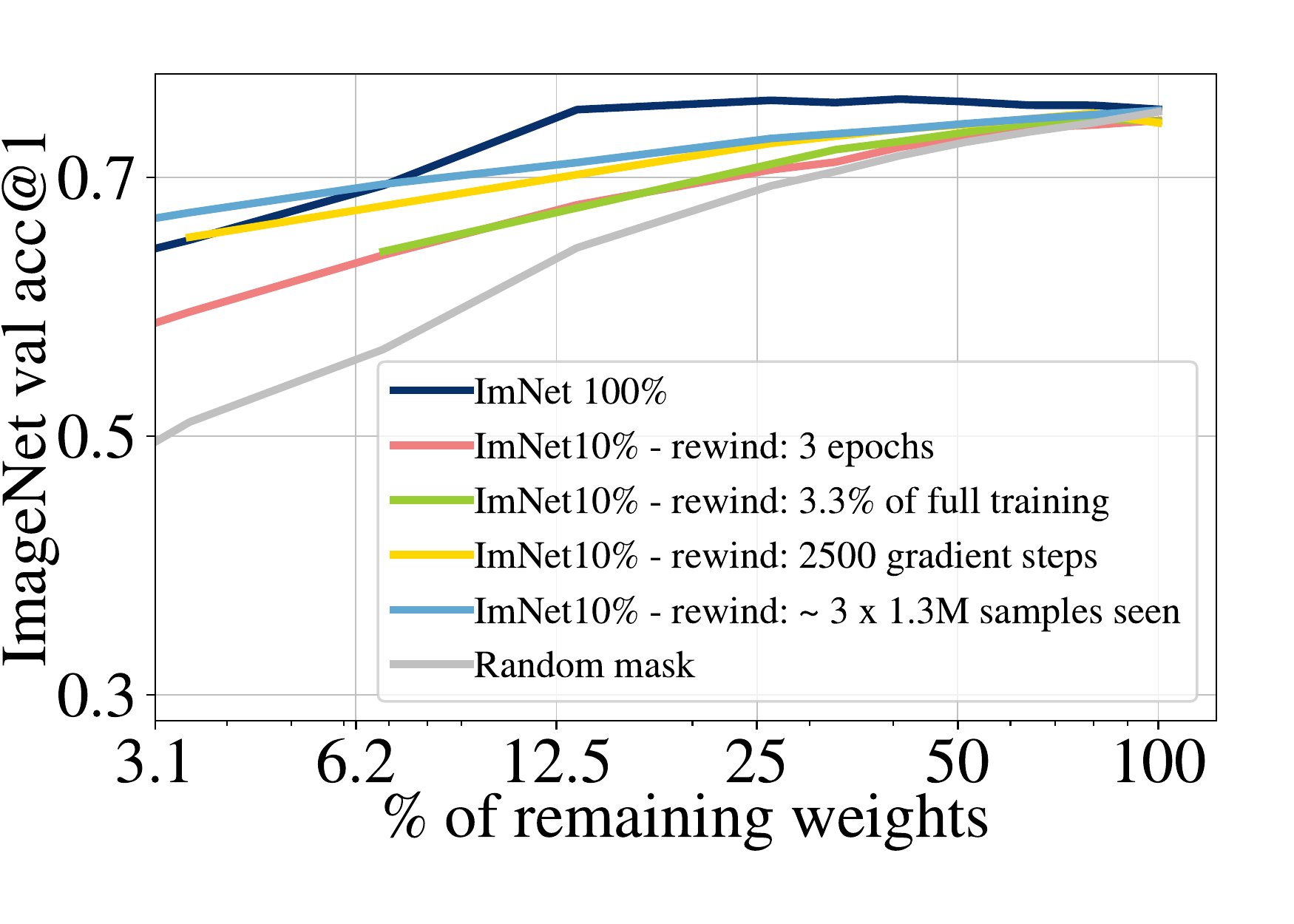}
\caption{
ImageNet top $1$ validation accuracy of winning tickets initialization found with a subset of $10\%$ of ImageNet dataset.
We show the influence of different values for the late resetting parameter.
}
\label{fig:late}
\end{figure}

We follow \cite{frankle2019lottery} and use late resetting (or \textit{rewind}) for the winning tickets generation process.
Indeed, before re-training a winning ticket, we reset its weights to their value ``early in training'' of the full over-parameterized network.
In our work, we set the late resetting parameter to $1$ epoch on CIFAR-10.
However, when dataset size, total number of epochs, mini-batch sizes or learning rate vary, it becomes more complicated to choose a rewind criterion that guarantees a fair comparison between all settings.
A choice can be to rewind at a point where ``the same amount of information'' has been processed.
Thus, in our work, we choose to set the rewind parameter to $3 \times 1,280,000$ samples for all our experiments on ImageNet, which corresponds to $3$ epochs on full ImageNet.
We describe in Table~\ref{comp} to what this rewind parameter corresponds to in terms of number of epoch, number of data samples seen, number of gradient update and percentage of total training for our different experiments.
Moreover, we show in Figure~\ref{fig:late} the performance of winning tickets generated using $10\%$ of ImageNet with different values of rewind.
Each of the considered value corresponds to keeping one of the criteria (number of epoch, number of data samples seen, number of gradient update or percentage of total training) fixed compared to the rewind parameter on full ImageNet (first row of Table~\ref{comp}).

\begin{table}[h]
  \centering
  \setlength{\tabcolsep}{1pt}
  \begin{tabular}{lcccccccc}
    \toprule
&& \# epochs && \# samples seen && \# grad updates && \% training \\
    \midrule
ImNet 100\% && $3$ && $\sim 3 \times 1,280,000$ && $2500$ && $3.3\%$ \\
    \midrule
ImNet 10\% && $30$ && $\sim 3 \times 1,280,000$ && $5000$ && $15\%$ \\
    \bottomrule
  \end{tabular}
\caption{}
\label{comp}
\end{table}

\subsection{More architectures for evaluating self-supervised masks}

\begin{figure*}[t]
\centering
\subfloat[Random Reinit]{
\includegraphics[width=0.24\linewidth]{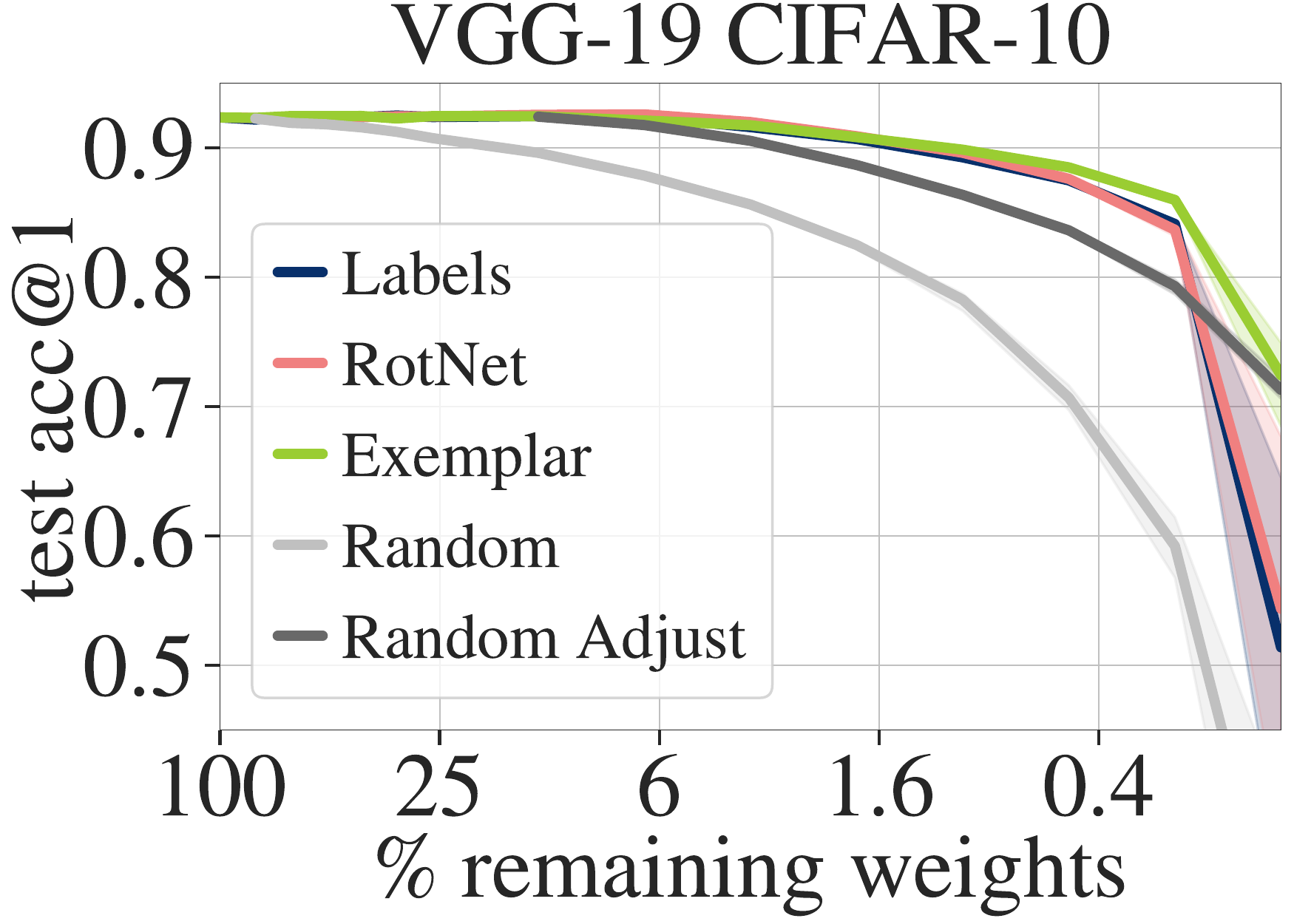}
\includegraphics[width=0.24\linewidth]{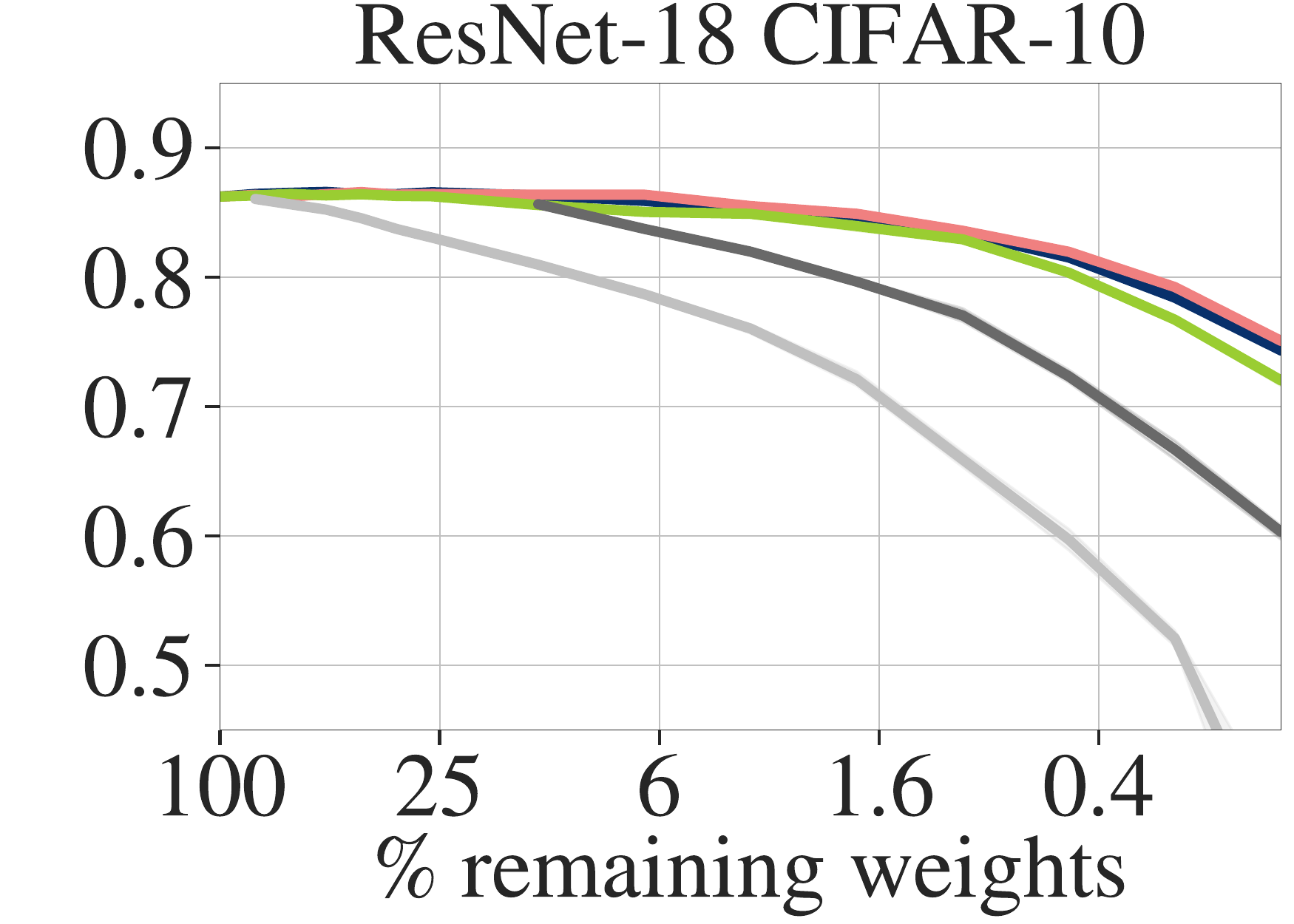}
\includegraphics[width=0.24\linewidth]{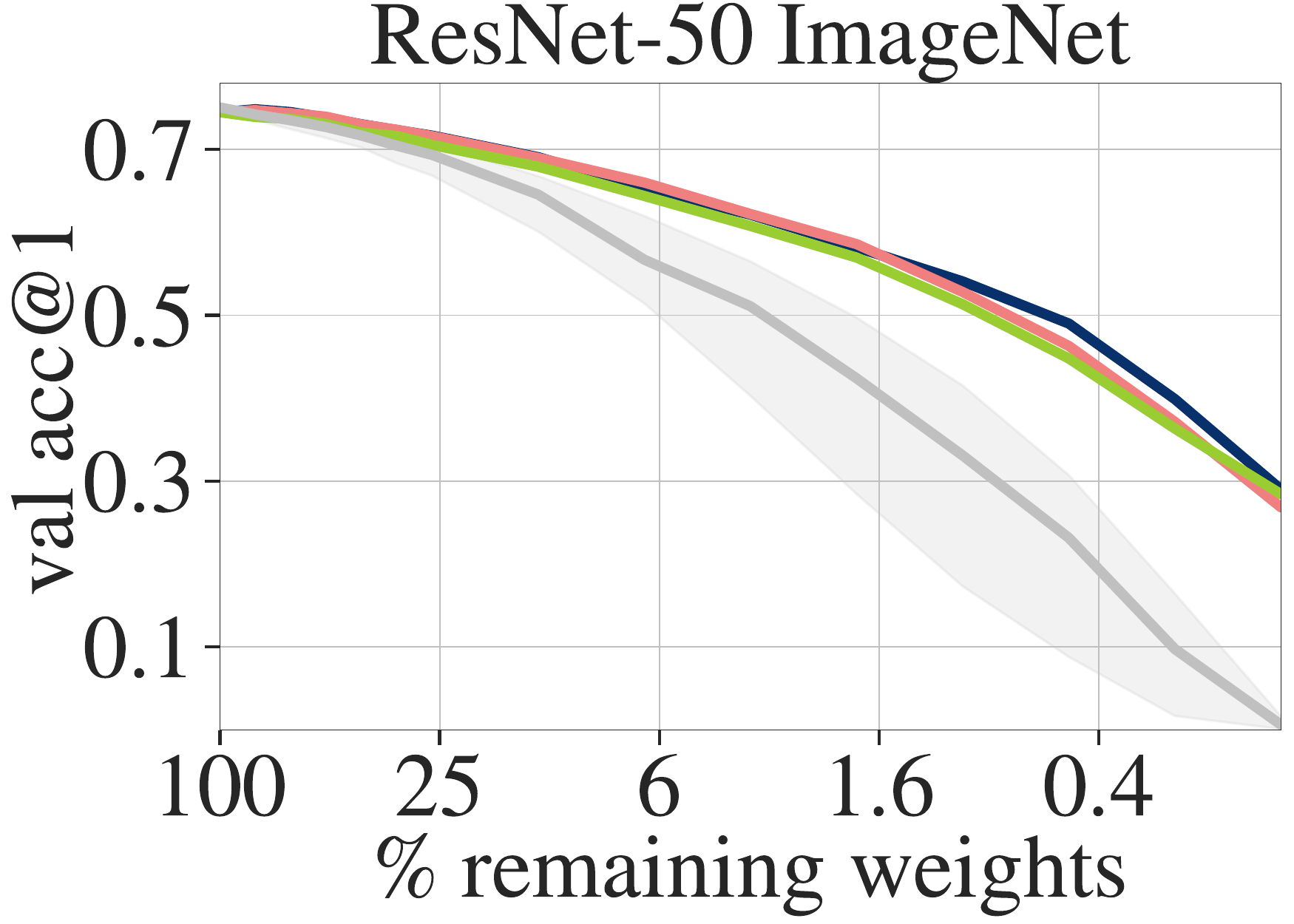}
\includegraphics[width=0.24\linewidth]{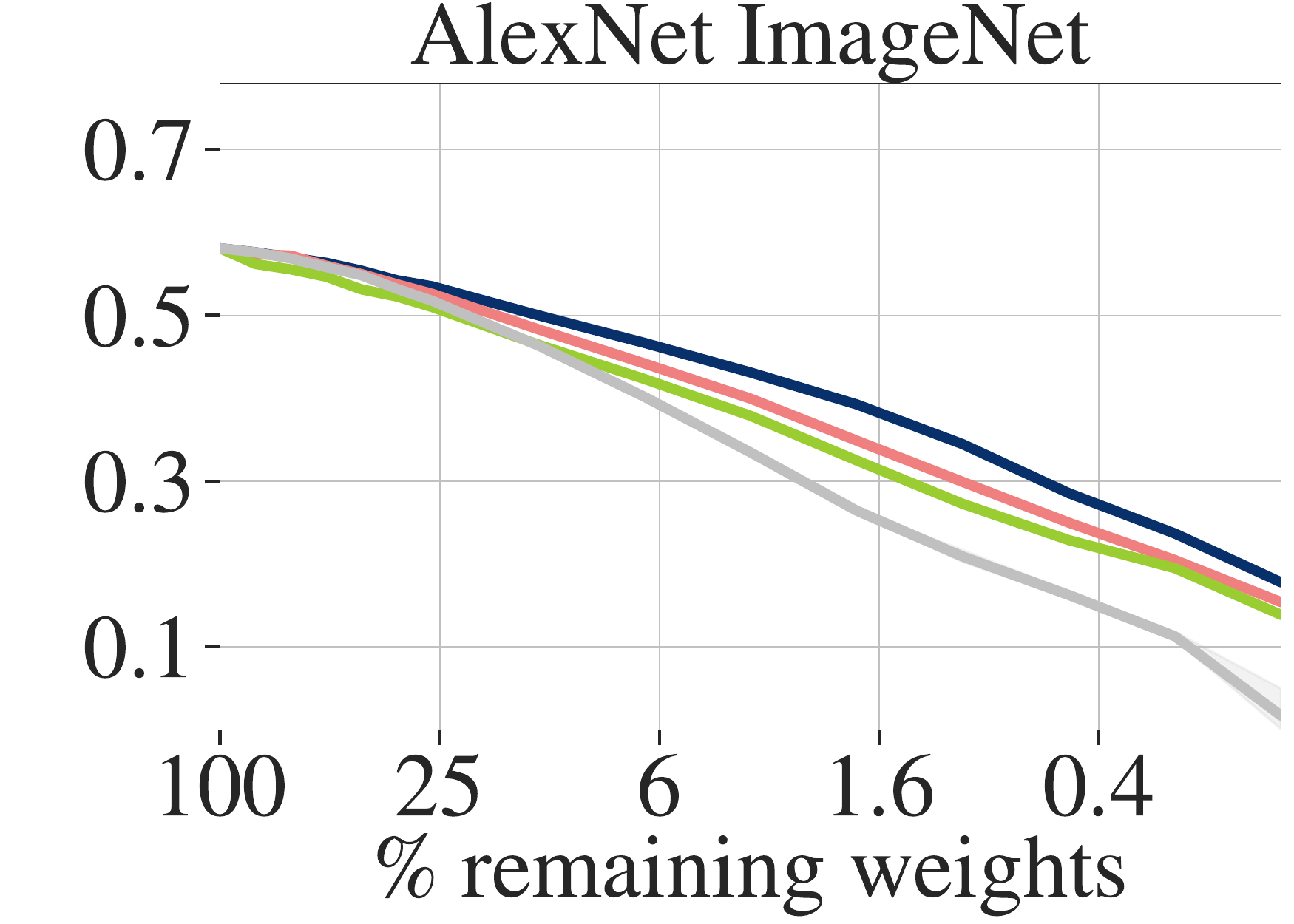}
\label{fig:apliu}
}

\subfloat[Winning Tickets]{
\includegraphics[width=0.24\linewidth]{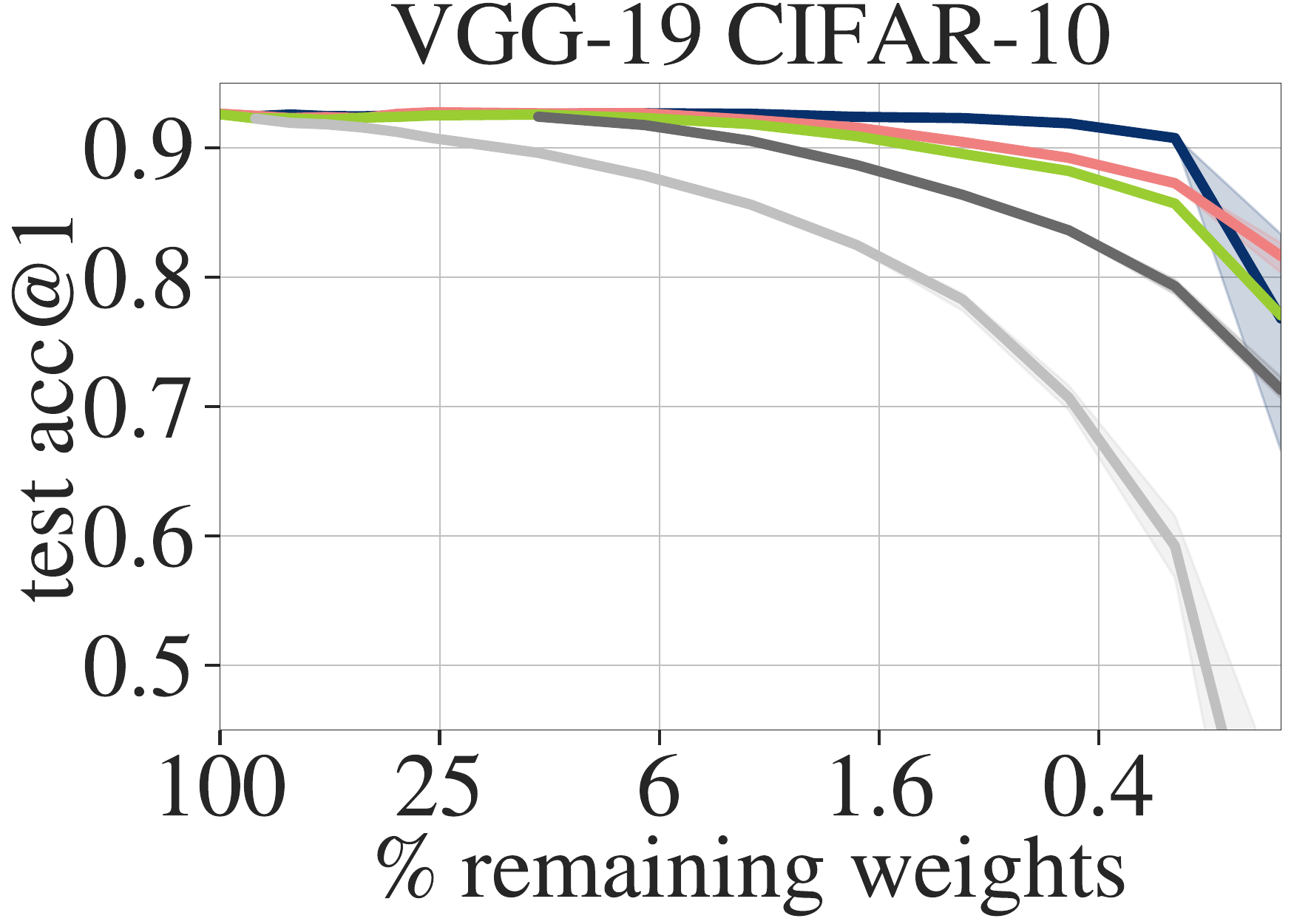}
\includegraphics[width=0.24\linewidth]{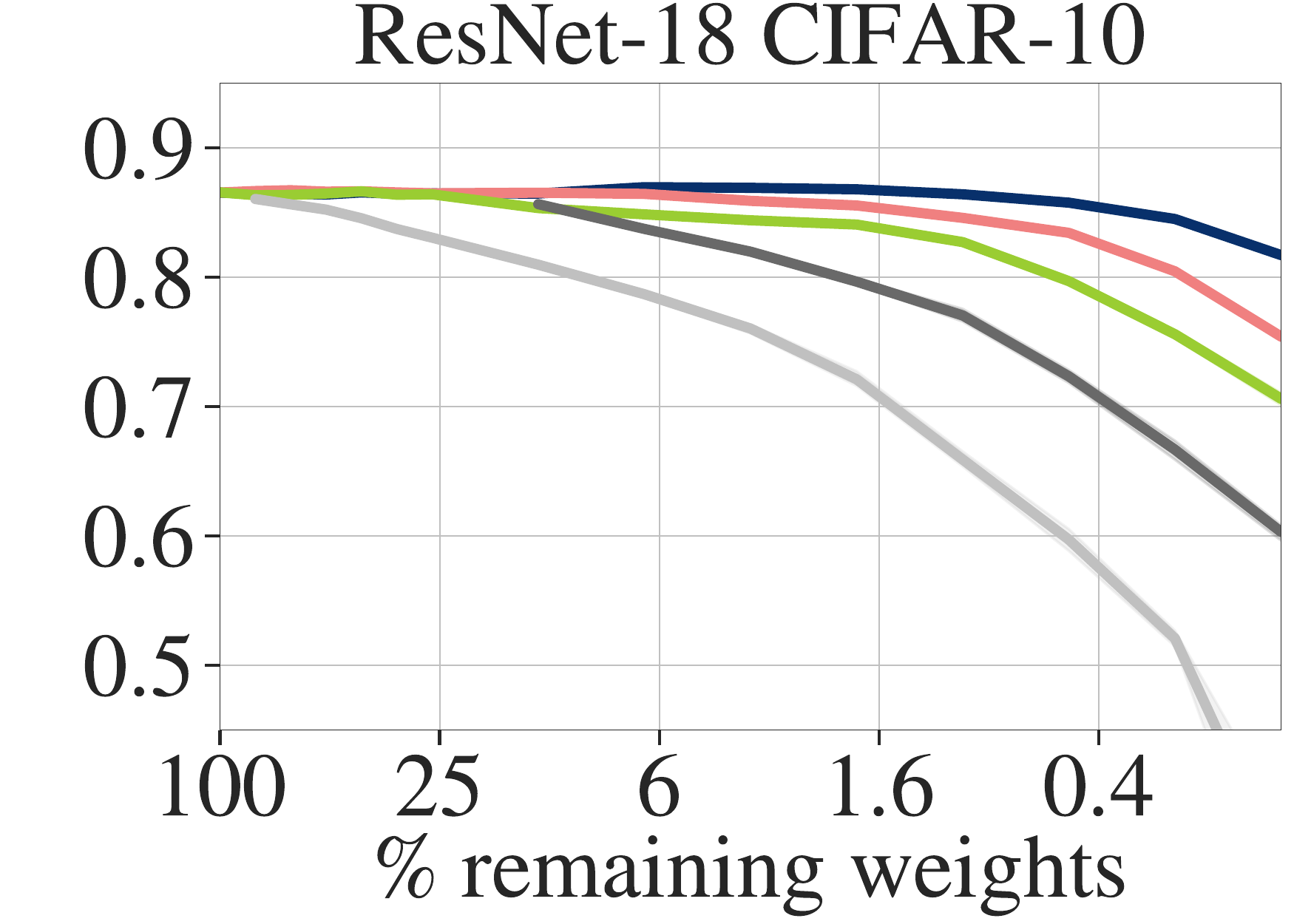}
\includegraphics[width=0.24\linewidth]{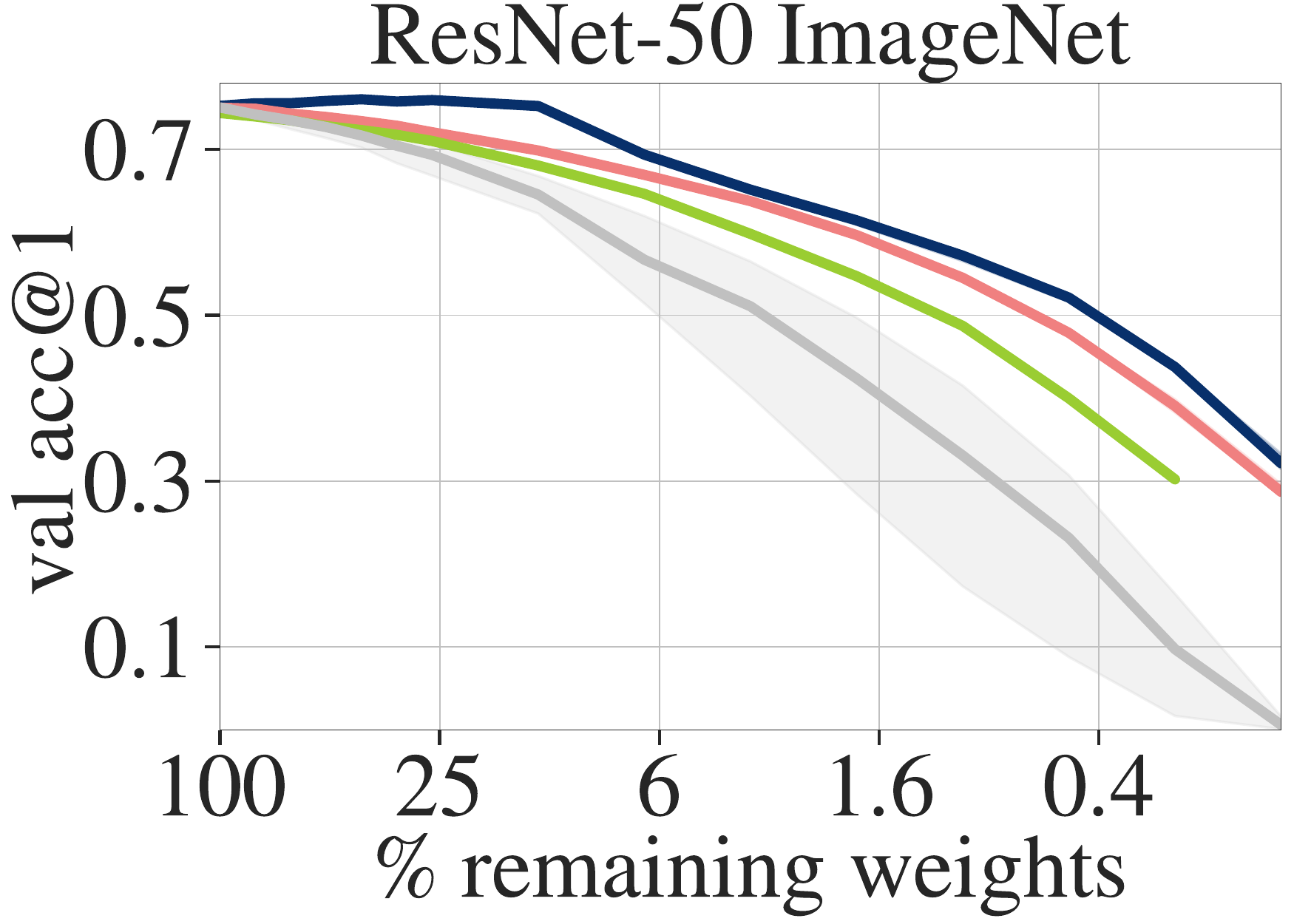}
\includegraphics[width=0.24\linewidth]{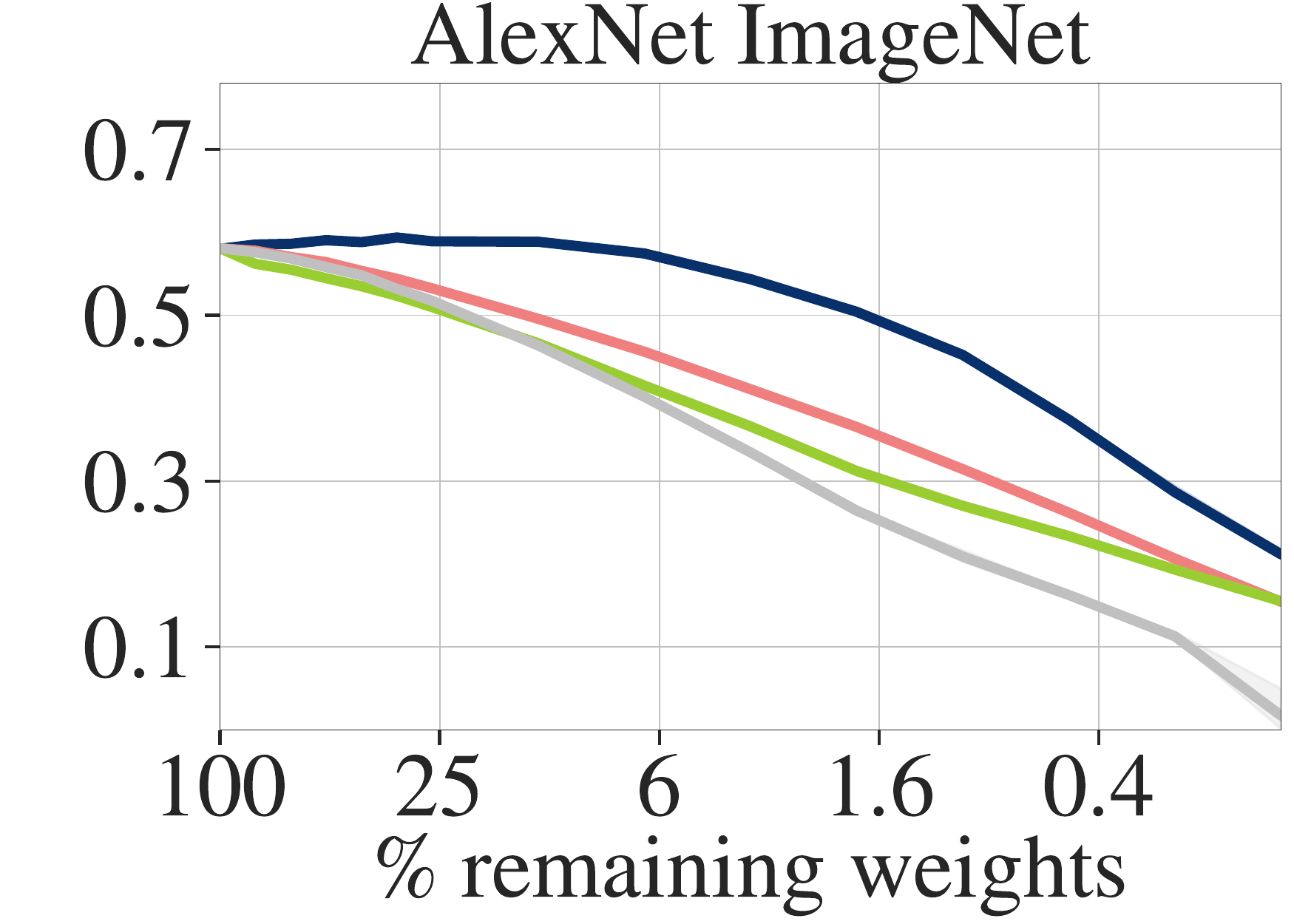}
\label{fig:apwt}
}
\caption{
We report CIFAR-10 test (left columns) and ImageNet val (right columns) top-$1$ accuracy for subnetworks pruned without labels, i.e. with self-supervised tasks: RotNet or Exemplar.
The x-axis corresponds to different pruning ratios.
We use two different schemes for weight initialization:
Random Reinit~\protect\subref{fig:apliu}: random re-initialization of the subnetwork~\cite{liu2018rethinking};
Winning Tickets~\protect\subref{fig:apwt}: inherited from early phase of training like the lottery tickets hypothesis~\cite{frankle2019lottery}.
Architectures are VGG-$19$ and ResNet-$18$ for CIFAR-$10$ and ResNet-$50$ and AlexNet for ImageNet.
We compare the performance with standard supervised pruning (dark blue) and random (grey) subnetworks which consist of randomly permuted masks and randomly drawn weights from the initialization distribution.
On CIFAR-10, deep models are highly sparse with only $\sim15\%$ of non-zero weights.
Thus, we adjust the random baseline to start with the correct mask at the natural level of network sparsity.
}

\label{fig:transfer}
\end{figure*}

\noindent\textbf{Re-training self-supervised masks on label classification.}
More results for this experiment can be found in Figure~\ref{fig:transfer}.
\\

\begin{figure*}[t]
\centering
  \begin{tabular}{cccc}
\includegraphics[width=0.22\linewidth]{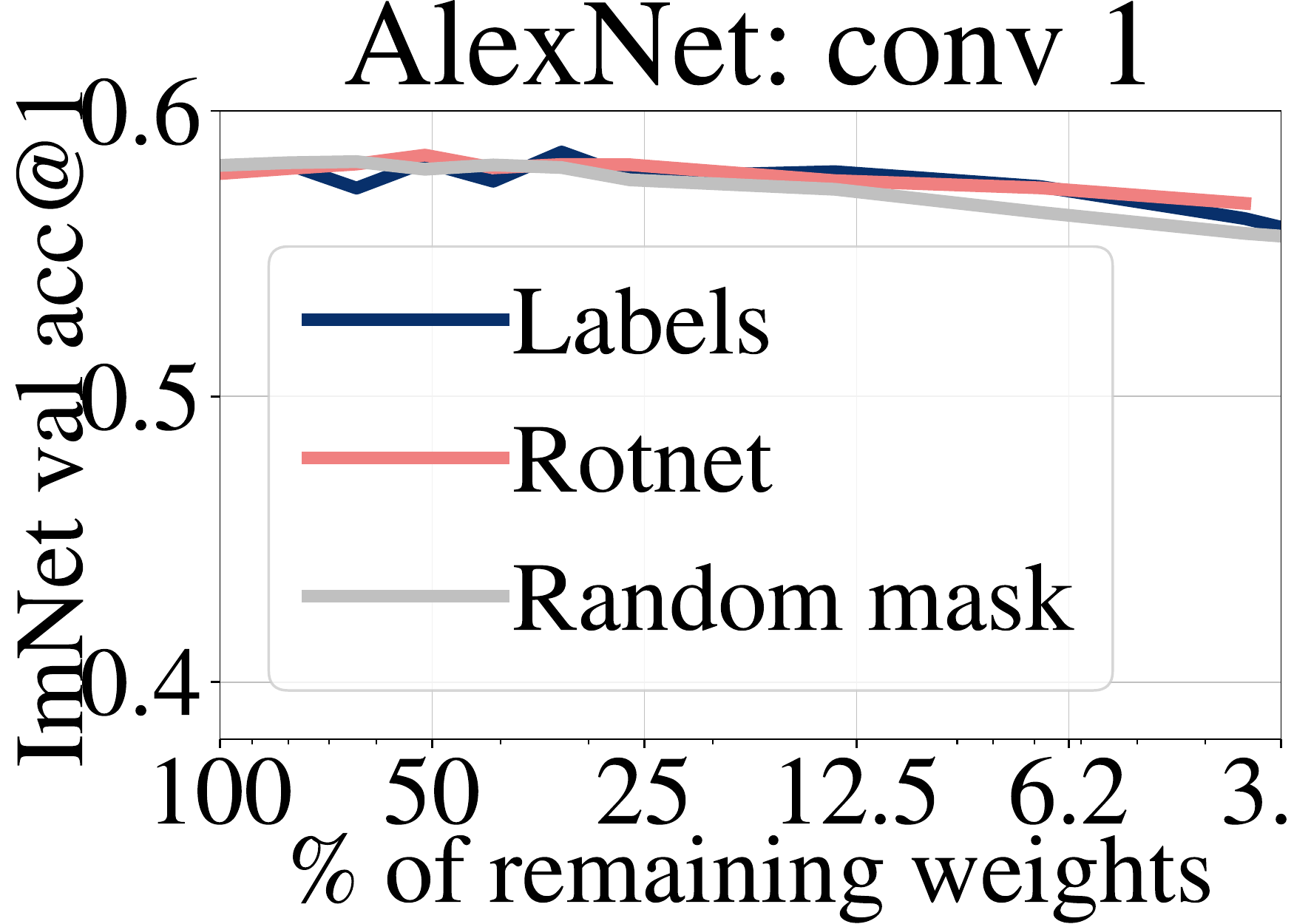} &
\includegraphics[width=0.22\linewidth]{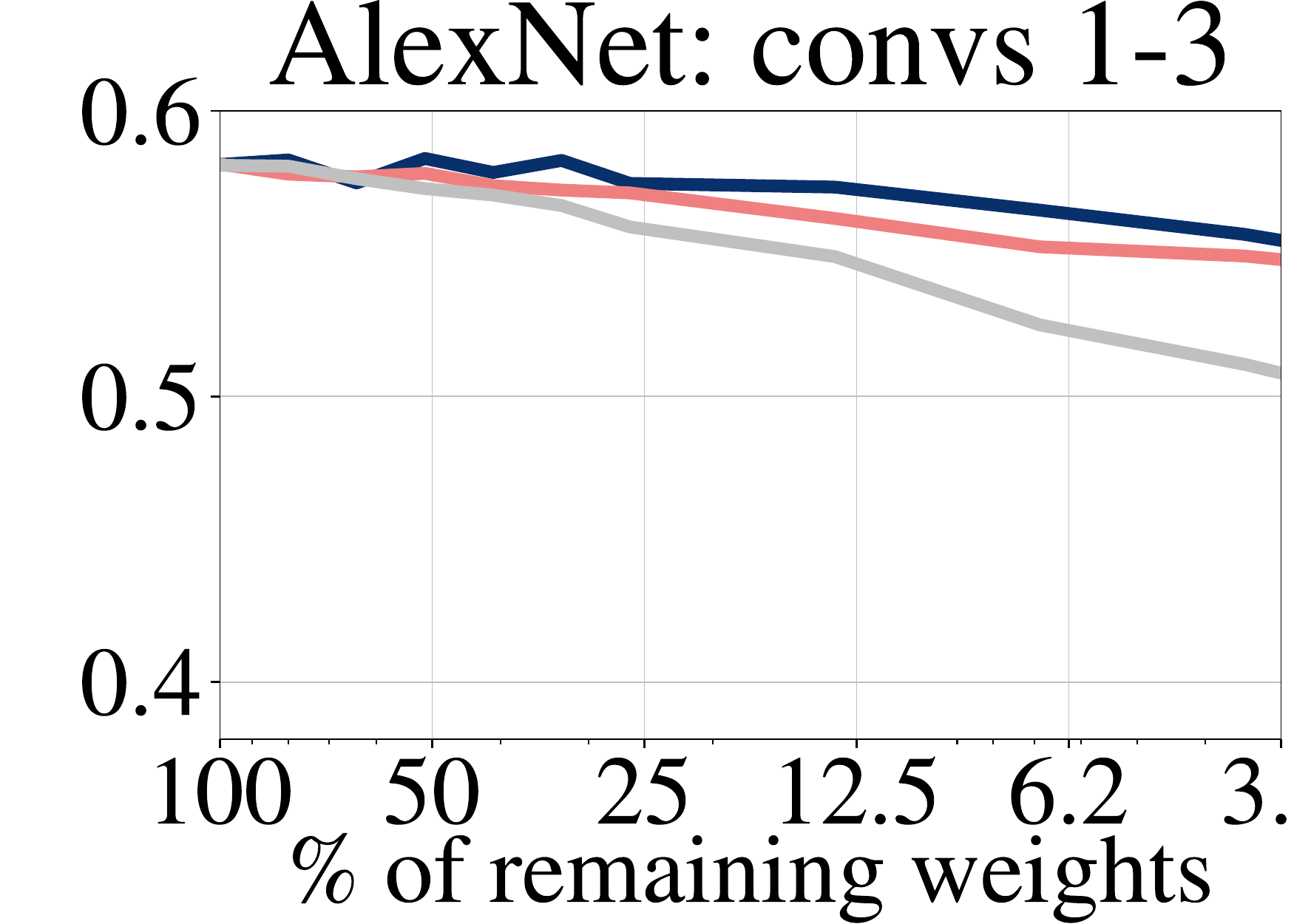} &
\includegraphics[width=0.22\linewidth]{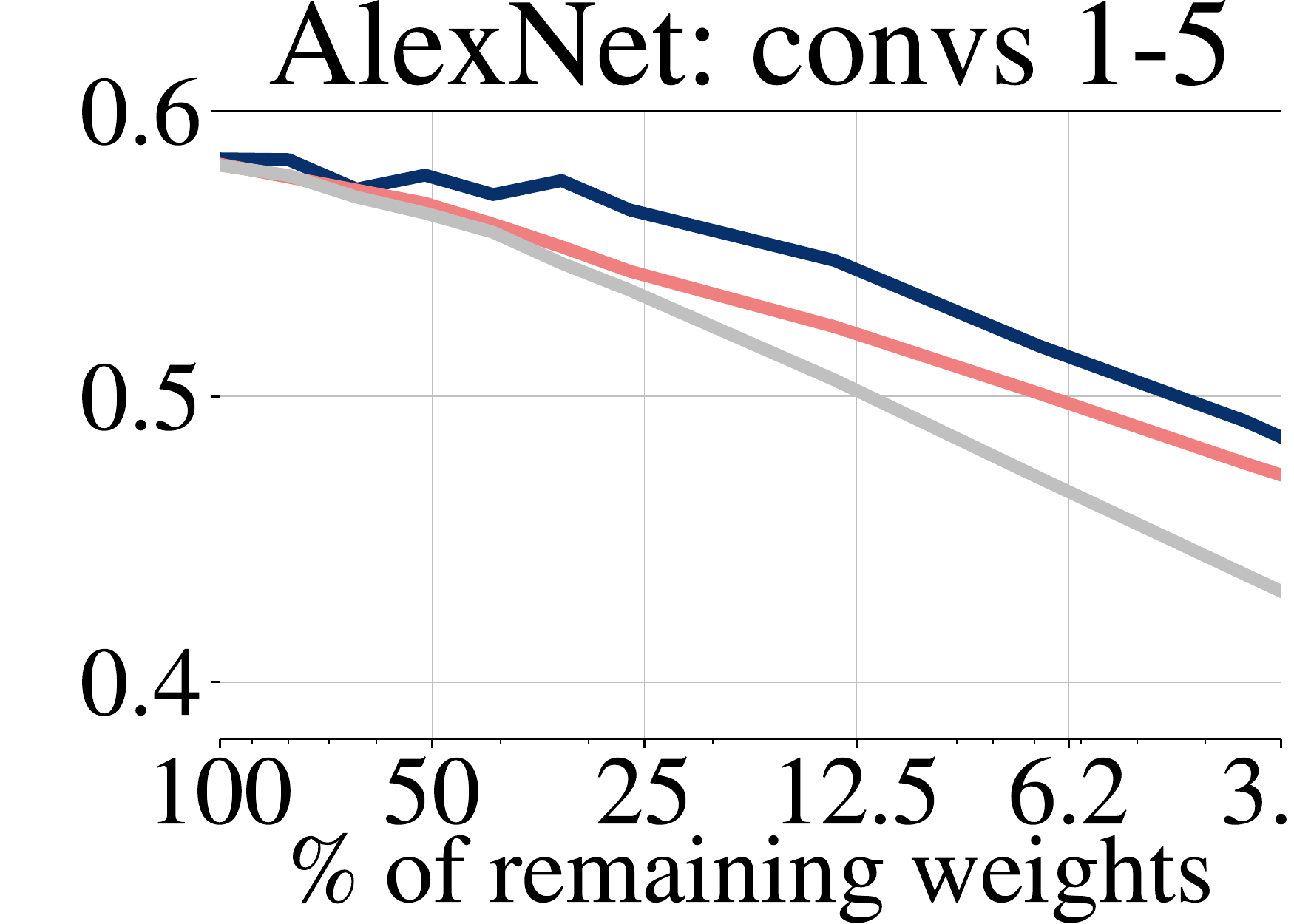} &
\includegraphics[width=0.22\linewidth]{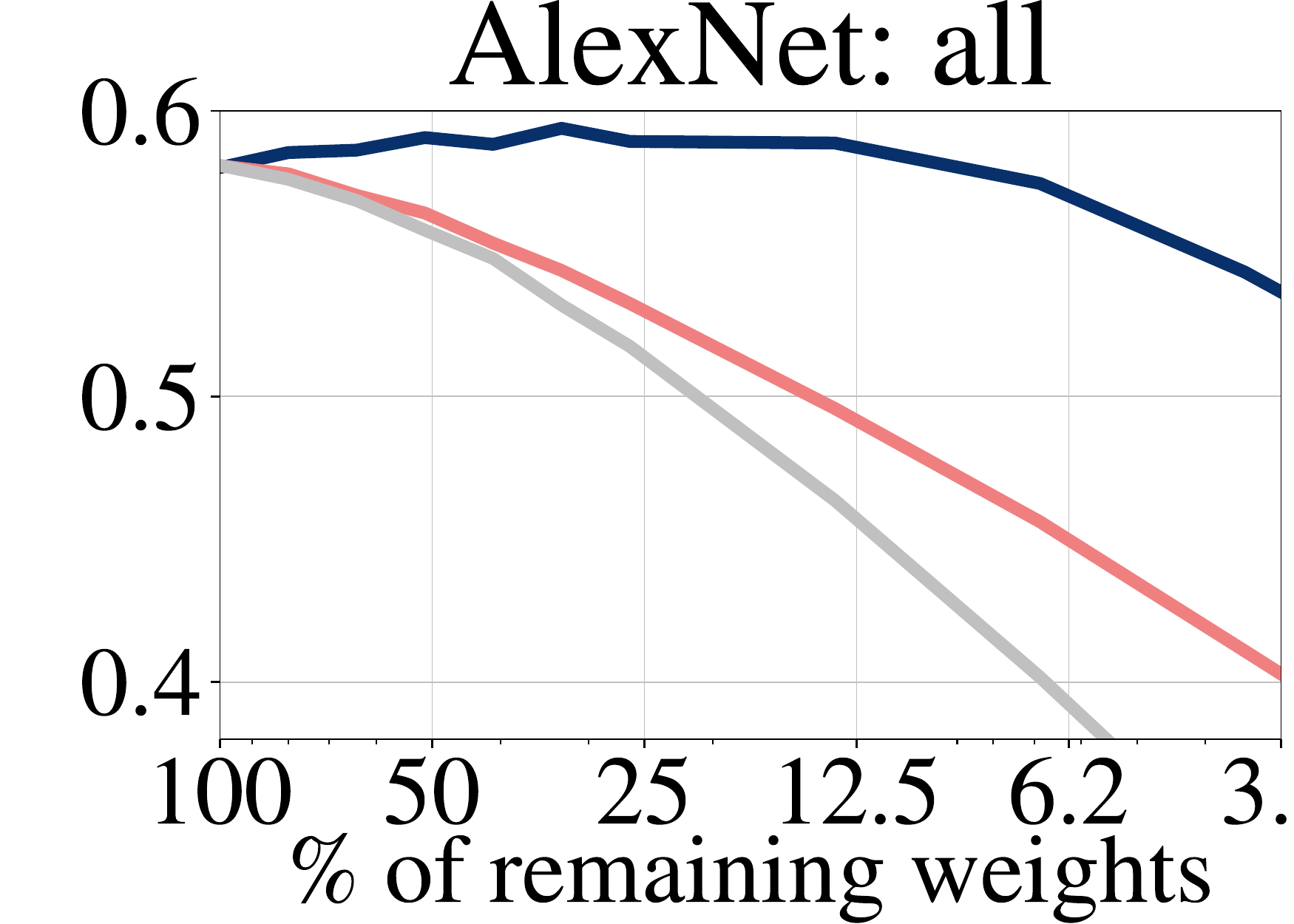} \\
  \end{tabular}
\caption{
ImageNet top $1$ validation accuracy of winning tickets generated by pruning partly or entirely (all) a network with $2$ generation tasks: labels classification or RotNet for AlexNet.
We also show for reference results when the network layers are randomly pruned.
}
\label{ap:layers}
\end{figure*}

\noindent\textbf{Layerwise winning tickets.}
We show more results about the layerwise winning tickets experiment.
We generate winning tickets by pruning only the $n$ first convolutional layers of a network, and leaving the remaining of the network unpruned.
On AlexNet, we consider $4$ situations.
From left to right in Figure~\ref{ap:layers}), we prune the first convolutiona layer; up to the third convolutiona layer; up to the fifth convolutional layer; or the whole network.
In Figure~\ref{ap:layers}, we verify that for the convolutional layers the gap of performance between labels winning tickets initializations and self-supervised ones remains narrow, while it becomes much wider when pruning the MLP.
Note that, even if this effect is particularly visible with an AlexNet, it is happening with most self-supervisedly trained network as can be seen in the main paper.

\subsection{Sparse trained networks on CIFAR-10}
\begin{figure}[t]
\centering
  \begin{tabular}{cc}
\includegraphics[width=0.48\linewidth]{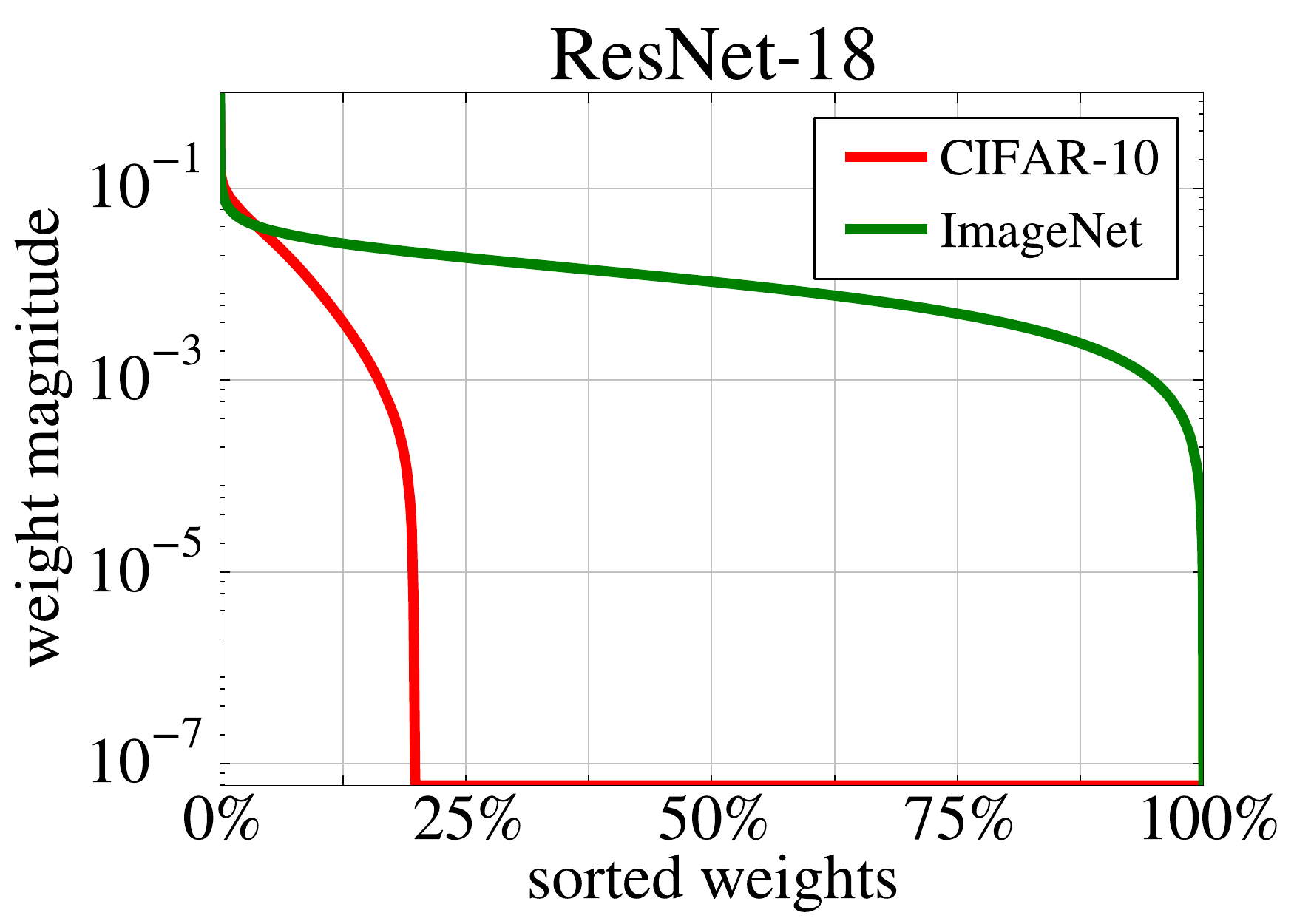} &
\includegraphics[width=0.48\linewidth]{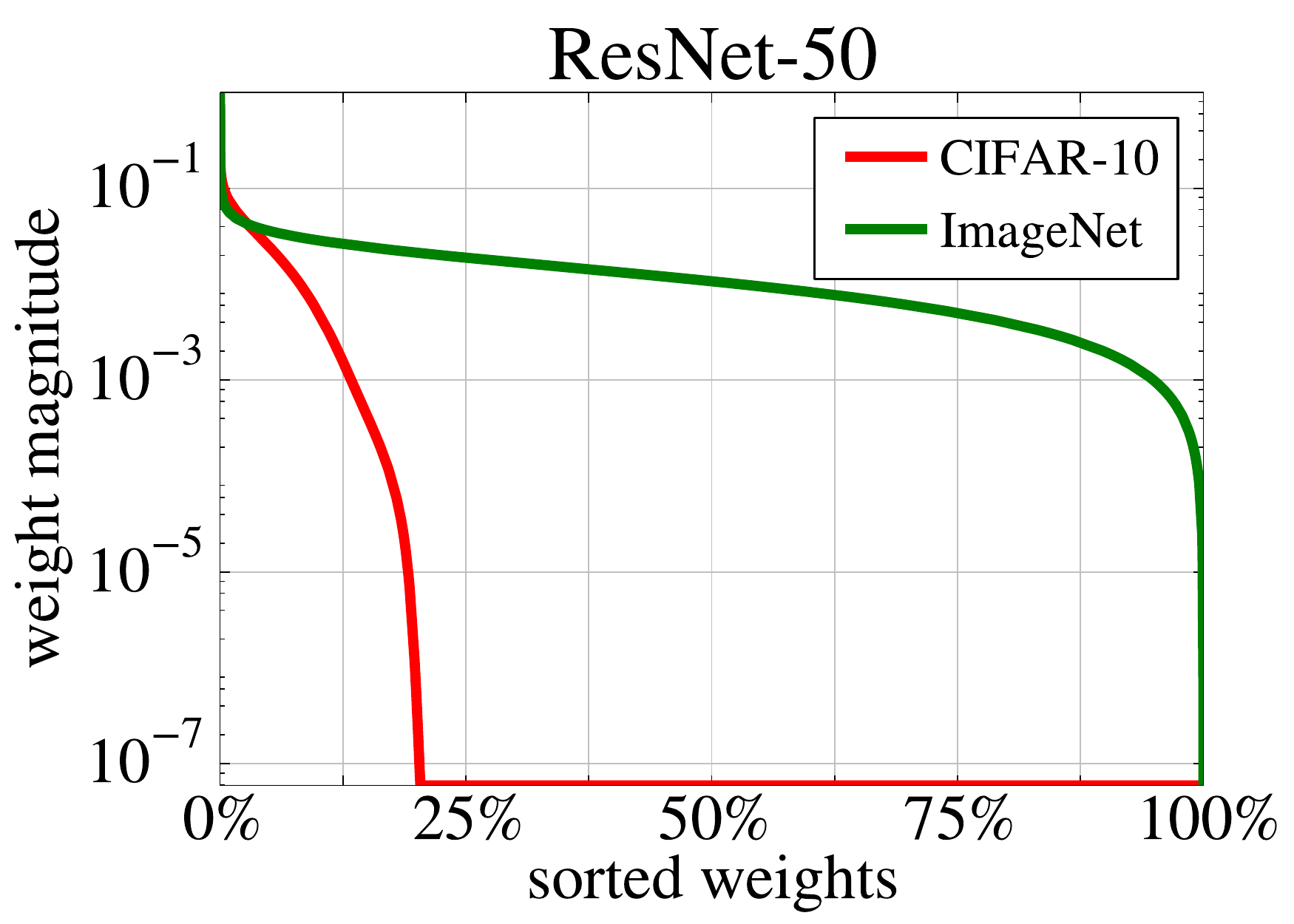} \\
	  \includegraphics[width=0.48\linewidth]{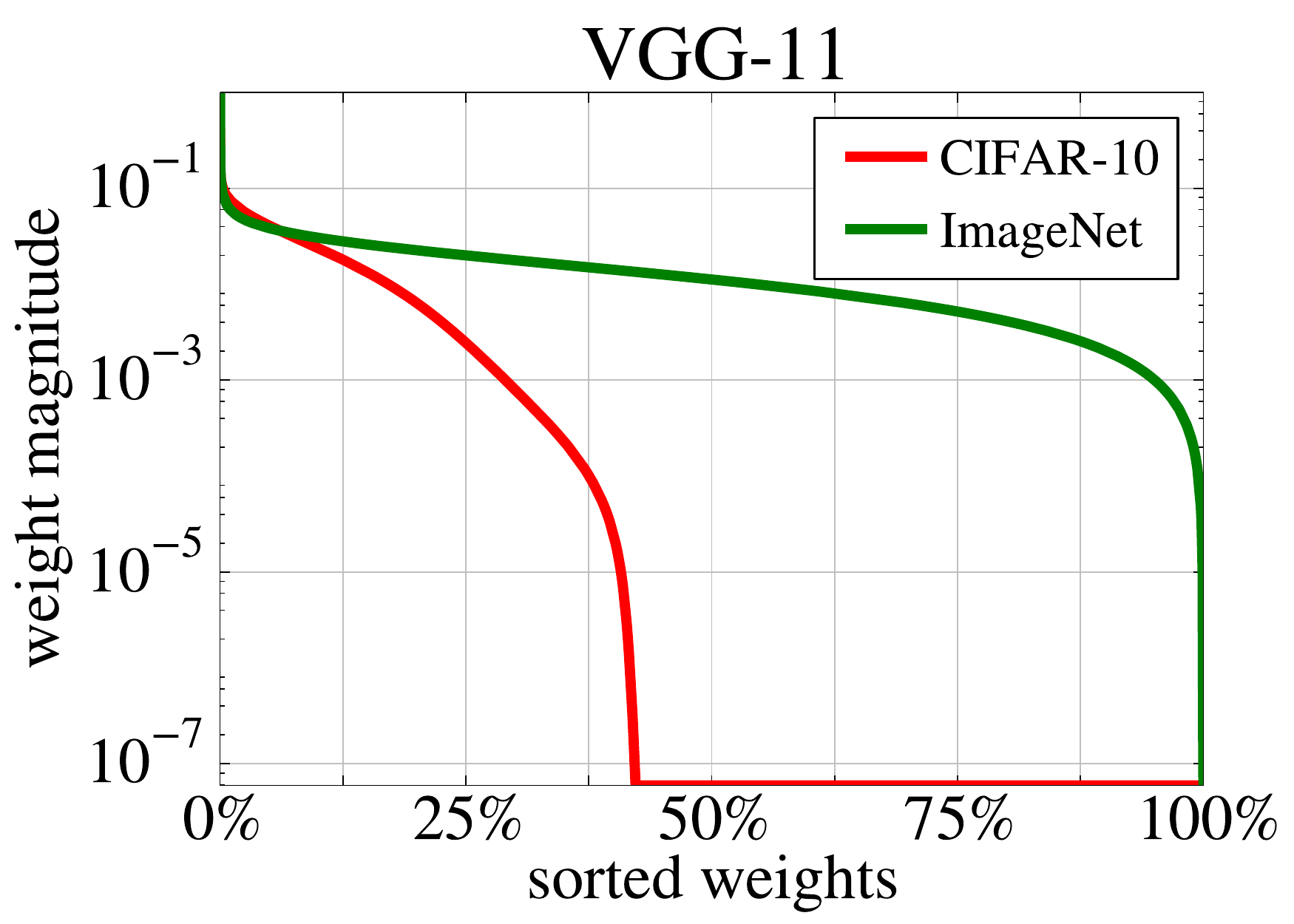} &
\includegraphics[width=0.48\linewidth]{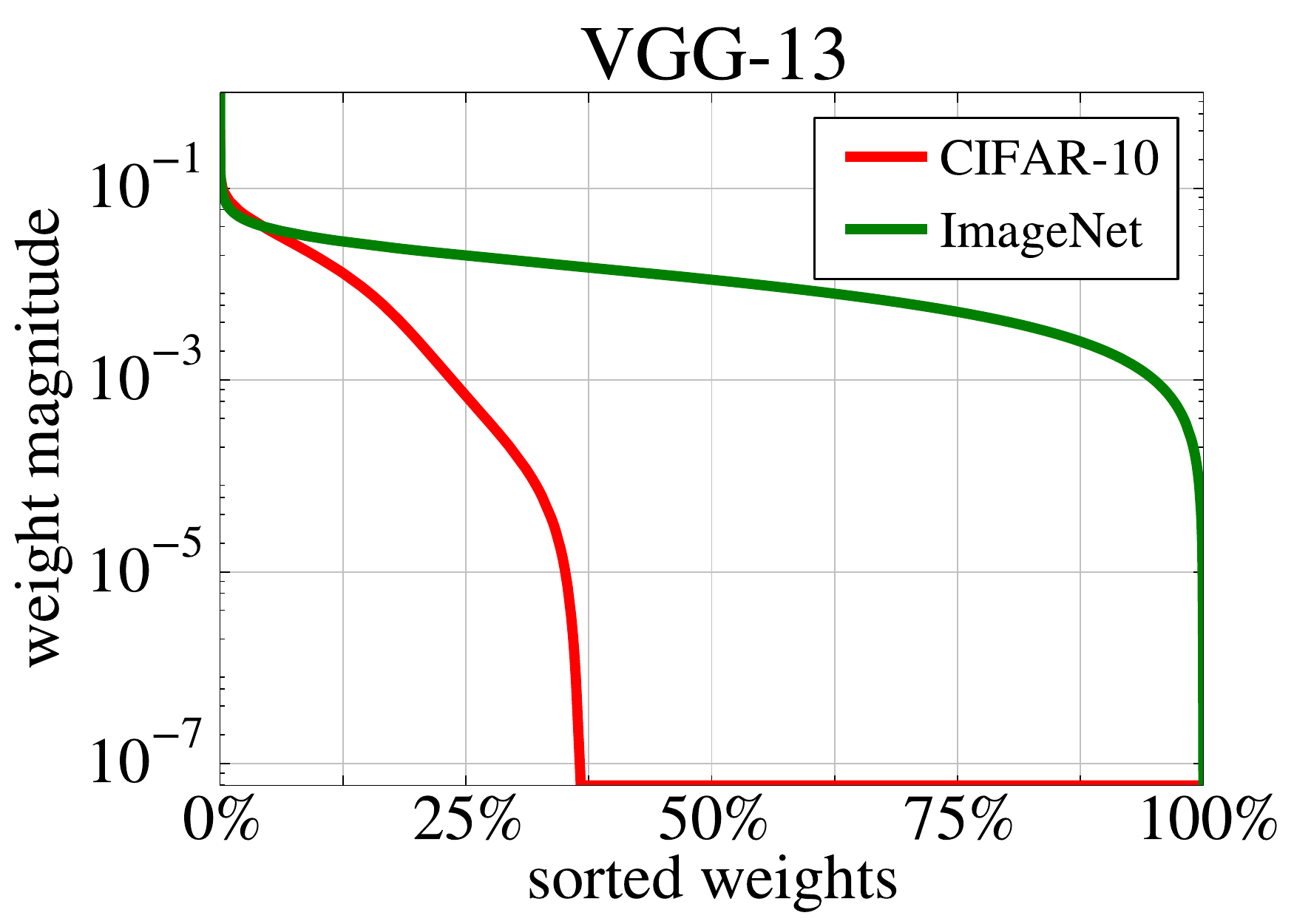} \\
	  \includegraphics[width=0.48\linewidth]{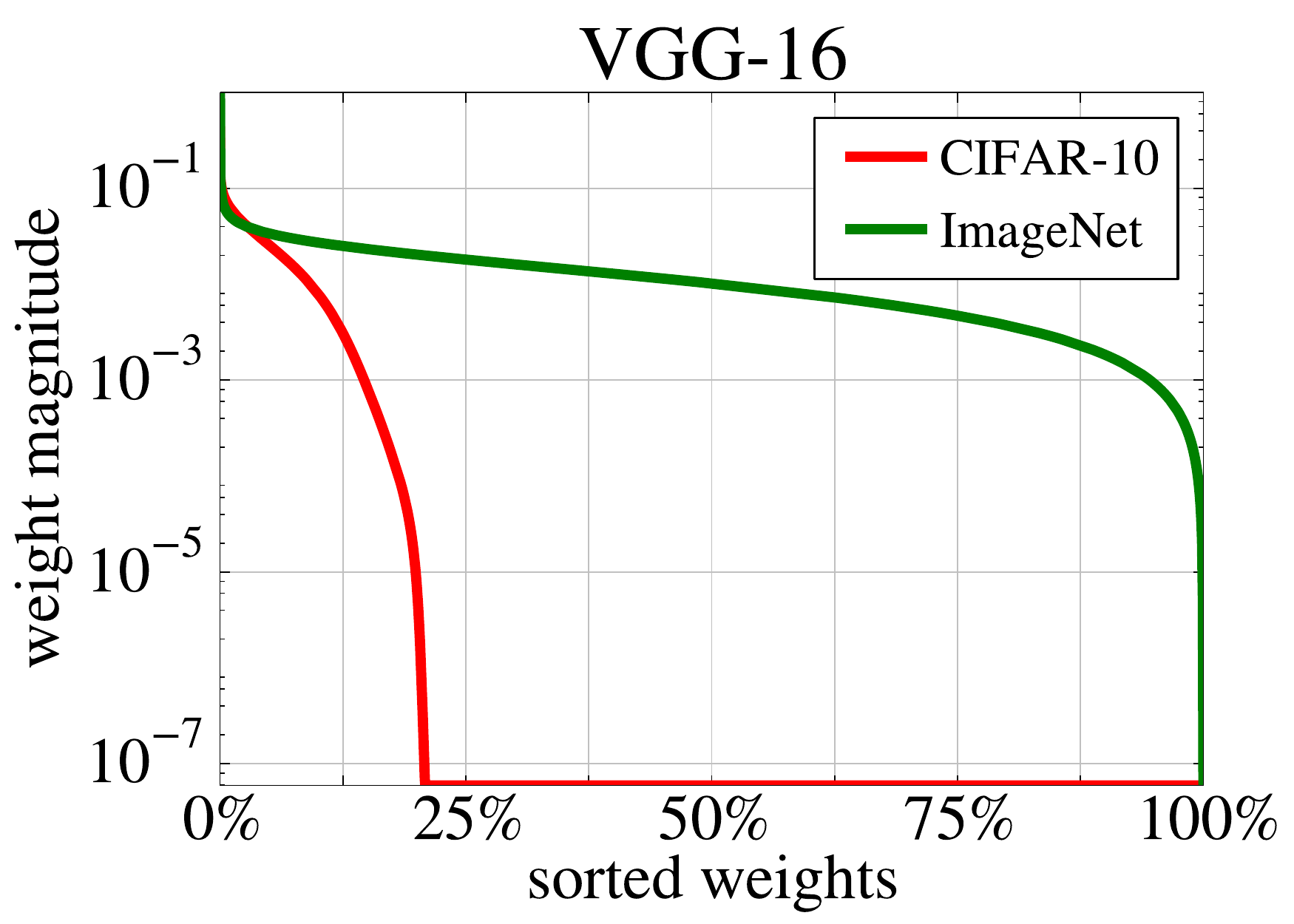} &
\includegraphics[width=0.48\linewidth]{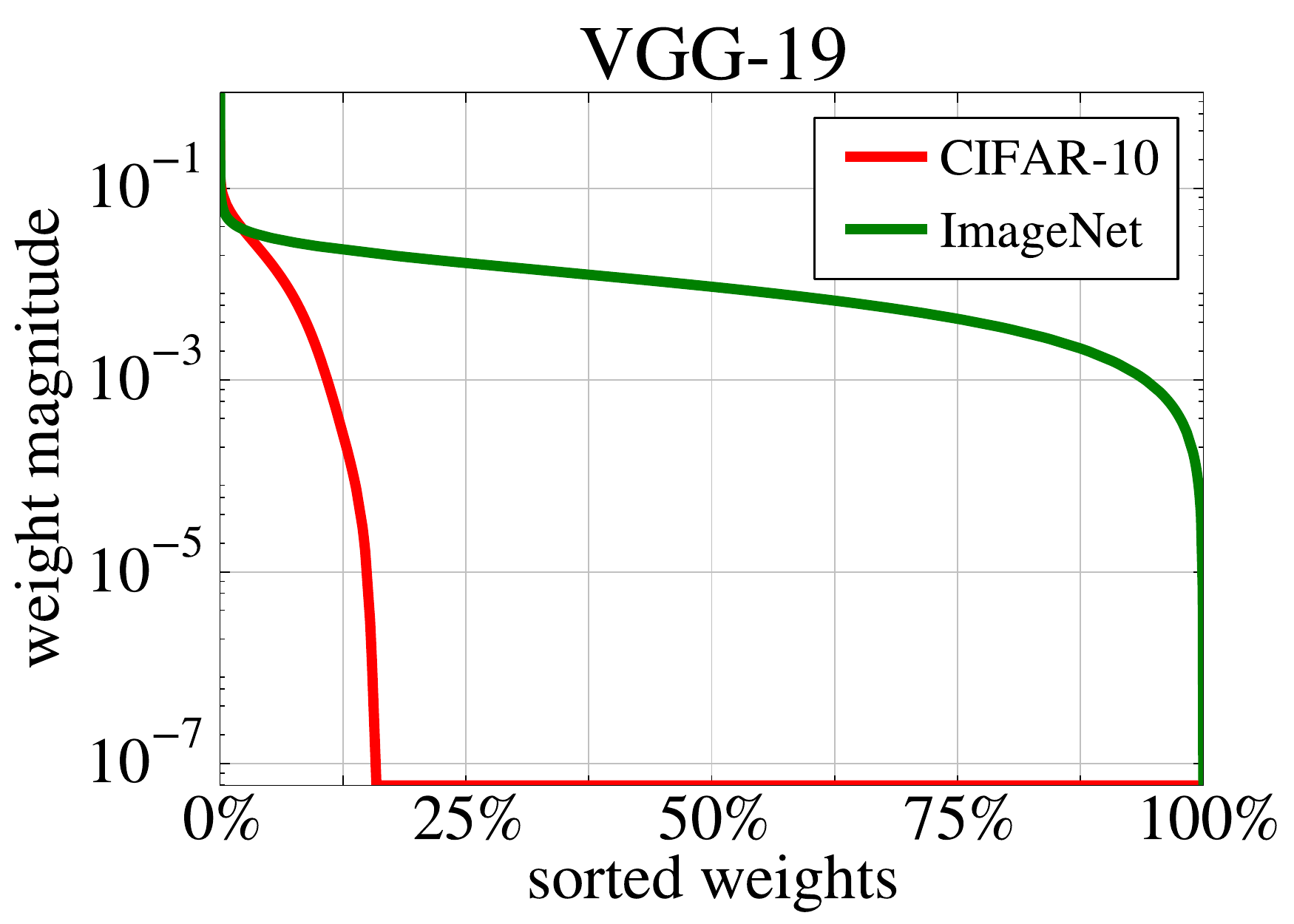} \\
  \end{tabular}
\caption{
Magnitude of the weights of a trained network on two different datasets: CIFAR-10 (green) and ImageNet (red) with different datasets.
We perform thresholding at machine precision value (bottom of y-axis).
}
\label{fig:manyratios}
\end{figure}

In this appendix, in Figure~\ref{fig:manyratios}, we provide results on more architectures about the proportion of weights zeroed during training on CIFAR-10 compared to ImageNet.
Note that we do not consider the batch-norm layers parameters, nor the parameters of the last fully-connected layer (since we do not prune it in our setting).
For all the considered VGGs we use the modified version of~\cite{morcos2019one}, replacing the final MLP by a fully connected layer.

\subsection{Hyperparameters and model details}

We detail in this appendix the different hyperparameters used in our experiments.
We use Adam optimizer on CIFAR-10.
On ImageNet, unless specified otherwise, we perform standard data augmentation consisting in croppings of random sizes and aspect ratios and horizontal flips~\cite{krizhevsky2012imagenet}).
On CIFAR-10, we use horizontal flips and croppings of fixed size on a $2$-padded input image.

\begin{itemize}

\item \textbf{ImageNet Labels - full dataset - AlexNet}: we train for $90$ epochs with a total batch-size of $4096$ distributed over $8$ GPUs ($512$ samples per GPU), learning rate of $0.4$, weight decay of $0.0001$.
We decay the learning rate by a factor $10$ at epochs $30$ and $60$.

\item \textbf{ImageNet Labels - full dataset - ResNet-50}: we train for $90$ epochs with a total batch-size of $1536$ distributed over $16$ GPUs ($96$ samples per GPU), learning rate of $0.1$, weight decay of $0.0001$.
We decay the learning rate by a factor $10$ at epochs $50$, $65$ and $80$.

\item \textbf{ImageNet Labels - 10\% dataset}: we train for $200$ epochs with a total batch-size of $768$ distributed over $8$ GPUs ($96$ samples per GPU), learning rate of $0.1$ warmed up during the first $5$ epochs, weight decay of $0.001$.
We decay the learning rate by a factor $10$ at epochs $140$, $160$ and $180$.

\item \textbf{ImageNet Labels - 10\% of classes}: we train for $200$ epochs with a total batch-size of $768$ distributed over $8$ GPUs ($96$ samples per GPU), learning rate of $0.3$ warmed up during the first $5$ epochs, weight decay of $0.0003$.
We decay the learning rate by a factor $10$ at epochs $140$, $160$ and $180$.

\item \textbf{ImageNet Semi-supervised RotNet}: We reproduce the semi-supervised method of \cite{zhai2019s} and follow precisely their hyperparameter.
We train for $200$ epochs with a total batch-size of $2048$ distributed over $32$ GPUs ($64$ samples per GPU), learning rate of $0.1$ warmed up during the first $5$ epochs, weight decay of $0.0003$.
We decay the learning rate by a factor $10$ at epochs $140$, $160$ and $180$.

\item \textbf{ImageNet RotNet - ResNet-50}: we train for $90$ epochs with a total batch-size of $1536$ distributed over $16$ GPUs ($96$ samples per GPU), learning rate of $1$ warmed up during the first $5$ epochs, weight decay of $0.00001$.
We decay the learning rate by a factor $10$ at epochs $50$, $65$ and $80$

\item \textbf{ImageNet Exemplar - ResNet-50}: we train for $40$ epochs with a total batch-size of $1536$ distributed over $32$ GPUs ($48$ samples per GPU), learning rate of $0.3$, weight decay of $0.00001$.
We decay the learning rate by a factor $10$ at epochs $20$ and $30$.
We follow the Exemplar implementation based on triplet margin loss from~\cite{doersch2017multi}.
As the goal of the task is to learn invariance to data transformation, we use data color augmentation and random small rotations on top of standard data augmentation scheme.

\item \textbf{ImageNet NPID - ResNet-50}: we train for $150$ epochs with a total batch-size of $1024$ distributed over $8$ GPUs ($128$ samples per GPU), learning rate of $0.03$, weight decay of $0.0001$.
We use $16384$ negatives and a cosine annealing learning rate schedule~\cite{loshchilov2016sgdr}.

\item \textbf{ImageNet RotNet - AlexNet}: we train for $90$ epochs with a total batch-size of $8192$ distributed over $16$ GPUs ($512$ samples per GPU), learning rate of $0.5$ warmed up during the first $2$ epochs, weight decay of $0.00001$.
We decay the learning rate by a factor $10$ at epochs $30$ and $60$.

\item \textbf{ImageNet Exemplar - AlexNet}: we train for $30$ epochs with a total batch-size of $4096$ distributed over $16$ GPUs ($256$ samples per GPU), learning rate of $0.1$, weight decay of $0.0001$.
We decay the learning rate by a factor $10$ at epoch $20$.
We follow the Exemplar implementation based on triplet margin loss from \cite{doersch2017multi}.
We use data color augmentation and random small rotations on top of standard data augmentation scheme.

\item \textbf{CIFAR-10 Labels \& RotNet - VGG-19 \& ResNet-18}: we train for $160$ epochs with a total batch-size of $512$ on $1$ GPU, learning rate of $0.001$, weight decay of $0.0001$.
We decay the learning rate by a factor $10$ at epochs $80$ and $120$.

\item \textbf{CIFAR-10 Exemplar - VGG-19 \& ResNet-18}: we train for $180$ epochs with a total batch-size of $512$ on $1$ GPU, learning rate of $0.0003$, weight decay of $0.0001$.
We decay the learning rate by a factor $10$ at epochs $120$.
As the goal of the task is to learn invariance to data transformation, we use data color augmentation and random small rotations on top of standard data augmentation scheme.

\end{itemize}

%% file: paper.bbl
\begin{thebibliography}{10}
\providecommand{\url}[1]{#1}
\csname url@samestyle\endcsname
\providecommand{\newblock}{\relax}
\providecommand{\bibinfo}[2]{#2}
\providecommand{\BIBentrySTDinterwordspacing}{\spaceskip=0pt\relax}
\providecommand{\BIBentryALTinterwordstretchfactor}{4}
\providecommand{\BIBentryALTinterwordspacing}{\spaceskip=\fontdimen2\font plus
\BIBentryALTinterwordstretchfactor\fontdimen3\font minus
  \fontdimen4\font\relax}
\providecommand{\BIBforeignlanguage}[2]{{%
\expandafter\ifx\csname l@#1\endcsname\relax
\typeout{** WARNING: IEEEtran.bst: No hyphenation pattern has been}%
\typeout{** loaded for the language `#1'. Using the pattern for}%
\typeout{** the default language instead.}%
\else
\language=\csname l@#1\endcsname
\fi
#2}}
\providecommand{\BIBdecl}{\relax}
\BIBdecl

\bibitem{caron2019unsupervised}
M.~Caron, P.~Bojanowski, J.~Mairal, and A.~Joulin, ``Unsupervised pre-training
  of image features on non-curated data,'' in \emph{Proceedings of the
  International Conference on Computer Vision (ICCV)}, 2019.

\bibitem{goyal2019scaling}
P.~Goyal, D.~Mahajan, A.~Gupta, and I.~Misra, ``Scaling and benchmarking
  self-supervised visual representation learning,'' \emph{Proceedings of the
  International Conference on Computer Vision (ICCV)}, 2019.

\bibitem{bachman2019learning}
P.~Bachman, R.~D. Hjelm, and W.~Buchwalter, ``Learning representations by
  maximizing mutual information across views,'' \emph{arXiv preprint
  arXiv:1906.00910}, 2019.

\bibitem{henaff2019data}
O.~J. H{\'e}naff, A.~Razavi, C.~Doersch, S.~Eslami, and A.~v.~d. Oord,
  ``Data-efficient image recognition with contrastive predictive coding,''
  \emph{arXiv preprint arXiv:1905.09272}, 2019.

\bibitem{he2016deep}
K.~He, X.~Zhang, S.~Ren, and J.~Sun, ``Deep residual learning for image
  recognition,'' in \emph{Proceedings of the Conference on Computer Vision and
  Pattern Recognition (CVPR)}, 2016.

\bibitem{lecun1990optimal}
Y.~LeCun, J.~S. Denker, and S.~A. Solla, ``Optimal brain damage,'' in
  \emph{Advances in Neural Information Processing Systems (NIPS)}, 1990.

\bibitem{louizos2017learning}
C.~Louizos, M.~Welling, and D.~P. Kingma, ``Learning sparse neural networks
  through $ l\_0 $ regularization,'' in \emph{International Conference on
  Learning Representations (ICLR)}, 2018.

\bibitem{han2015learning}
S.~Han, J.~Pool, J.~Tran, and W.~Dally, ``Learning both weights and connections
  for efficient neural network,'' in \emph{Advances in Neural Information
  Processing Systems (NIPS)}, 2015.

\bibitem{noroozi2016unsupervised}
M.~Noroozi and P.~Favaro, ``Unsupervised learning of visual representations by
  solving jigsaw puzzles,'' in \emph{Proceedings of the European Conference on
  Computer Vision (ECCV)}, 2016.

\bibitem{gidaris2018unsupervised}
S.~Gidaris, P.~Singh, and N.~Komodakis, ``Unsupervised representation learning
  by predicting image rotations,'' in \emph{International Conference on
  Learning Representations (ICLR)}, 2018.

\bibitem{frankle2018lottery}
J.~Frankle and M.~Carbin, ``The lottery ticket hypothesis: Finding sparse,
  trainable neural networks,'' in \emph{International Conference on Learning
  Representations (ICLR)}, 2019.

\bibitem{frankle2019lottery}
J.~Frankle, G.~K. Dziugaite, D.~M. Roy, and M.~Carbin, ``Linear mode
  connectivity and the lottery ticket hypothesis,'' \emph{arXiv preprint
  arXiv:1912.05671}, 2019.

\bibitem{liu2018rethinking}
Z.~Liu, M.~Sun, T.~Zhou, G.~Huang, and T.~Darrell, ``Rethinking the value of
  network pruning,'' in \emph{International Conference on Learning
  Representations (ICLR)}, 2019.

\bibitem{morcos2019one}
A.~S. Morcos, H.~Yu, M.~Paganini, and Y.~Tian, ``One ticket to win them all:
  generalizing lottery ticket initializations across datasets and optimizers,''
  \emph{Advances in Neural Information Processing Systems (NeurIPS)}, 2019.

\bibitem{li2016pruning}
H.~Li, A.~Kadav, I.~Durdanovic, H.~Samet, and H.~P. Graf, ``Pruning filters for
  efficient convnets,'' in \emph{International Conference on Learning
  Representations (ICLR)}, 2017.

\bibitem{prakash2019repr}
A.~Prakash, J.~Storer, D.~Florencio, and C.~Zhang, ``Repr: Improved training of
  convolutional filters,'' in \emph{Proceedings of the Conference on Computer
  Vision and Pattern Recognition (CVPR)}, 2019.

\bibitem{guo2016dynamic}
Y.~Guo, A.~Yao, and Y.~Chen, ``Dynamic network surgery for efficient dnns,'' in
  \emph{Advances in Neural Information Processing Systems (NIPS)}, 2016.

\bibitem{dong2017learning}
X.~Dong, S.~Chen, and S.~Pan, ``Learning to prune deep neural networks via
  layer-wise optimal brain surgeon,'' in \emph{Advances in Neural Information
  Processing Systems (NIPS)}, 2017.

\bibitem{ullrich2017soft}
K.~Ullrich, E.~Meeds, and M.~Welling, ``Soft weight-sharing for neural network
  compression,'' in \emph{International Conference on Learning Representations
  (ICLR)}, 2017.

\bibitem{rastegari2016xnor}
M.~Rastegari, V.~Ordonez, J.~Redmon, and A.~Farhadi, ``Xnor-net: Imagenet
  classification using binary convolutional neural networks,'' in
  \emph{Proceedings of the European Conference on Computer Vision (ECCV)},
  2016.

\bibitem{chen2015compressing}
W.~Chen, J.~Wilson, S.~Tyree, K.~Weinberger, and Y.~Chen, ``Compressing neural
  networks with the hashing trick,'' in \emph{Proceedings of the International
  Conference on Machine Learning (ICML)}, 2015.

\bibitem{zhou2019deconstructing}
H.~Zhou, J.~Lan, R.~Liu, and J.~Yosinski, ``Deconstructing lottery tickets:
  Zeros, signs, and the supermask,'' in \emph{"Workshop on Identifying and
  Understanding Deep Learning Phenomena (ICML)"}, 2019.

\bibitem{yu2019playing}
H.~Yu, S.~Edunov, Y.~Tian, and A.~S. Morcos, ``Playing the lottery with rewards
  and multiple languages: lottery tickets in {RL} and {NLP},''
  \emph{International Conference on Learning Representations (ICLR)}, 2020.

\bibitem{doersch2015unsupervised}
C.~Doersch, A.~Gupta, and A.~A. Efros, ``Unsupervised visual representation
  learning by context prediction,'' in \emph{Proceedings of the International
  Conference on Computer Vision (ICCV)}, 2015.

\bibitem{wang2015unsupervised}
X.~Wang and A.~Gupta, ``Unsupervised learning of visual representations using
  videos,'' in \emph{Proceedings of the International Conference on Computer
  Vision (ICCV)}, 2015.

\bibitem{zhang2016colorful}
R.~Zhang, P.~Isola, and A.~A. Efros, ``Colorful image colorization,'' in
  \emph{Proceedings of the European Conference on Computer Vision (ECCV)},
  2016.

\bibitem{pathak2017learning}
D.~Pathak, R.~Girshick, P.~Doll{\'a}r, T.~Darrell, and B.~Hariharan, ``Learning
  features by watching objects move,'' in \emph{Proceedings of the Conference
  on Computer Vision and Pattern Recognition (CVPR)}, 2017.

\bibitem{tian2019contrastive}
Y.~Tian, D.~Krishnan, and P.~Isola, ``Contrastive multiview coding,''
  \emph{arXiv preprint arXiv:1906.05849}, 2019.

\bibitem{dosovitskiy2016discriminative}
A.~Dosovitskiy, P.~Fischer, J.~T. Springenberg, M.~Riedmiller, and T.~Brox,
  ``Discriminative unsupervised feature learning with exemplar convolutional
  neural networks,'' \emph{IEEE Transactions on Pattern Analysis and Machine
  Intelligence (TPAMI)}, 2016.

\bibitem{bojanowski2017unsupervised}
P.~Bojanowski and A.~Joulin, ``Unsupervised learning by predicting noise,'' in
  \emph{Proceedings of the International Conference on Machine Learning
  (ICML)}, 2017.

\bibitem{wu2018unsupervised}
Z.~Wu, Y.~Xiong, S.~X. Yu, and D.~Lin, ``Unsupervised feature learning via
  non-parametric instance discrimination,'' in \emph{Proceedings of the
  Conference on Computer Vision and Pattern Recognition (CVPR)}, 2018.

\bibitem{caron2018deep}
M.~Caron, P.~Bojanowski, A.~Joulin, and M.~Douze, ``Deep clustering for
  unsupervised learning of visual features,'' in \emph{Proceedings of the
  European Conference on Computer Vision (ECCV)}, 2018.

\bibitem{doersch2017multi}
C.~Doersch and A.~Zisserman, ``Multi-task self-supervised visual learning,'' in
  \emph{Proceedings of the International Conference on Computer Vision (ICCV)},
  2017.

\bibitem{deng2009imagenet}
J.~Deng, W.~Dong, R.~Socher, L.-J. Li, K.~Li, and L.~Fei-Fei, ``Imagenet: A
  large-scale hierarchical image database,'' in \emph{Proceedings of the
  Conference on Computer Vision and Pattern Recognition (CVPR)}, 2009.

\bibitem{krizhevsky2009learning}
A.~Krizhevsky, ``Learning multiple layers of features from tiny images,'' Tech.
  Rep., 2009.

\bibitem{krizhevsky2012imagenet}
A.~Krizhevsky, I.~Sutskever, and G.~E. Hinton, ``Imagenet classification with
  deep convolutional neural networks,'' in \emph{Advances in Neural Information
  Processing Systems (NIPS)}, 2012.

\bibitem{simonyan2014very}
K.~Simonyan and A.~Zisserman, ``Very deep convolutional networks for
  large-scale image recognition,'' \emph{arXiv preprint arXiv:1409.1556}, 2014.

\bibitem{paszke2017automatic}
A.~Paszke, S.~Gross, S.~Chintala, G.~Chanan, E.~Yang, Z.~DeVito, Z.~Lin,
  A.~Desmaison, L.~Antiga, and A.~Lerer, ``Automatic differentiation in
  pytorch,'' 2017.

\bibitem{everingham2010pascal}
M.~Everingham, L.~Van~Gool, C.~K. Williams, J.~Winn, and A.~Zisserman, ``The
  pascal visual object classes (voc) challenge,'' \emph{International Journal
  of Computer Vision (IJCV)}, 2010.

\bibitem{zhou2014learning}
B.~Zhou, A.~Lapedriza, J.~Xiao, A.~Torralba, and A.~Oliva, ``Learning deep
  features for scene recognition using places database,'' in \emph{Advances in
  Neural Information Processing Systems (NIPS)}, 2014.

\bibitem{jing2019self}
L.~Jing and Y.~Tian, ``Self-supervised visual feature learning with deep neural
  networks: A survey,'' \emph{IEEE Transactions on Pattern Analysis and Machine
  Intelligence (TPAMI)}, 2019.

\bibitem{zhai2019s}
X.~Zhai, A.~Oliver, A.~Kolesnikov, and L.~Beyer, ``S4l: Self-supervised
  semi-supervised learning,'' \emph{Proceedings of the International Conference
  on Computer Vision (ICCV)}, 2019.

\bibitem{miyato2018virtual}
T.~Miyato, S.-i. Maeda, M.~Koyama, and S.~Ishii, ``Virtual adversarial
  training: a regularization method for supervised and semi-supervised
  learning,'' \emph{IEEE Transactions on Pattern Analysis and Machine
  Intelligence (TPAMI)}, 2018.

\bibitem{lee2013pseudo}
D.-H. Lee, ``Pseudo-label: The simple and efficient semi-supervised learning
  method for deep neural networks,'' in \emph{Workshop on Challenges in
  Representation Learning, (ICML)}, 2013.

\bibitem{loshchilov2016sgdr}
I.~Loshchilov and F.~Hutter, ``Sgdr: Stochastic gradient descent with warm
  restarts,'' \emph{arXiv preprint arXiv:1608.03983}, 2016.

\end{thebibliography}
